\documentclass[10pt,twocolumn,letterpaper]{article}

\usepackage[pagenumbers]{cvpr} % To force page numbers, e.g. for an arXiv version

\usepackage{graphicx}
\usepackage{tabularx}
\usepackage{amsmath}
\usepackage{amssymb}
\usepackage{booktabs}
\usepackage{enumitem}
\usepackage{wrapfig}
\usepackage{etoolbox}
\BeforeBeginEnvironment{wrapfigure}{\setlength{\intextsep}{0pt}\setlength{\columnsep}{2pt}}
\usepackage[pagebackref,breaklinks,colorlinks]{hyperref}

\usepackage[capitalize]{cleveref}
\crefname{section}{Sec.}{Secs.}
\Crefname{section}{Section}{Sections}
\Crefname{table}{Table}{Tables}
\crefname{table}{Tab.}{Tabs.}

\newcommand{\p}{\mathrm{p}}
\newcommand{\co}{\mathbf{o}}
\newcommand{\cv}{\mathbf{v}}
\newcommand{\n}{\mathbf{n}}
\newcommand{\mi}{\mathbb{I}}
\newcommand{\ent}{\mathbb{H}}
\newcommand{\loss}{\mathcal{L}}
\newcommand{\jb}{\mathrm{J}}
\DeclareMathOperator*{\argmax}{arg\,max}

\newcommand{\A}{\mathbf{A}}
\newcommand{\B}{\mathbf{B}}

\def\cvprPaperID{768} % *** Enter the CVPR Paper ID here
\def\confName{CVPR}
\def\confYear{2023}

\let\oldtwocolumn\twocolumn
\renewcommand\twocolumn[1][]{%
    \oldtwocolumn[{#1}{
    \begin{center}
           \includegraphics[width=0.96\textwidth]{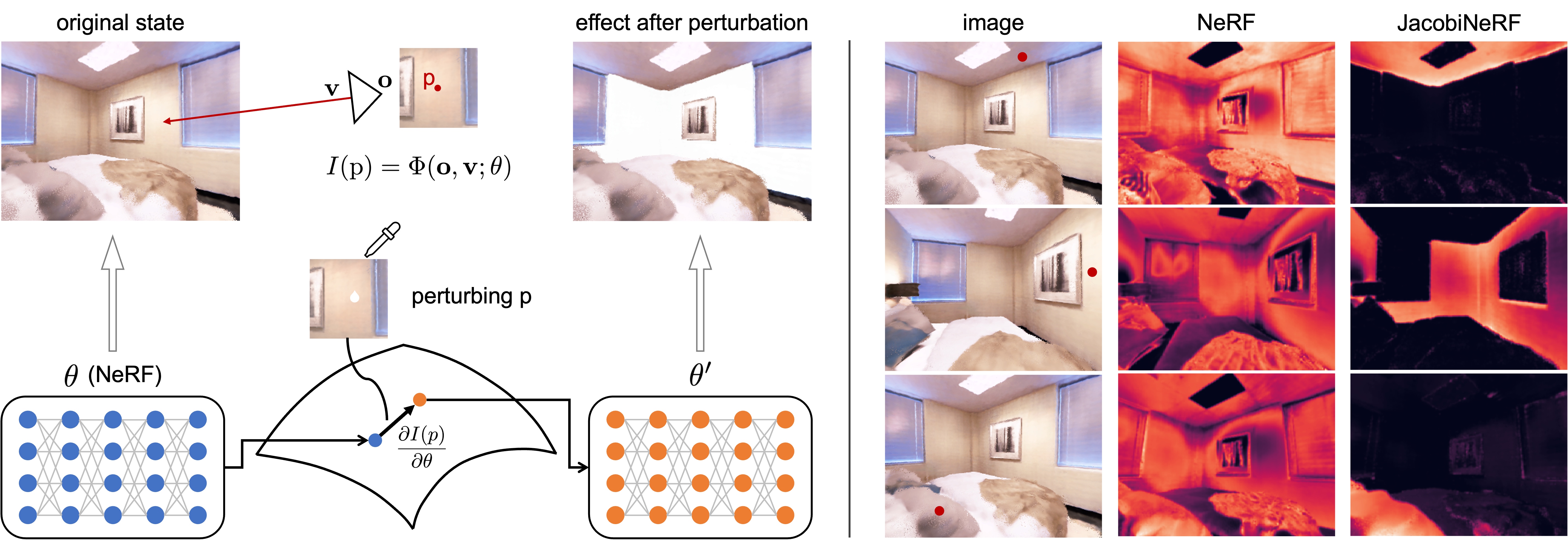}
           \captionof{figure}{\emph{Left}: a physical scene is composed of mutually-correlated entities, e.g., points belonging to the same part or object. We pursue a subspace of the parameters of an implicit scene representation so that when the scene is perturbed along the gradient of a specific point, a resonance emerges with other points having high mutual information. 
           For example, if the brightness of a point $\p$ on the wall changes, the other wall points should also change coherently (while points on the ceiling should be unaffected). \emph{Right}: a comparison between responses to perturbations along the gradient of the highlighted red point in the scene. The response from the original NeRF spreads all over the scene, while the localized response from the proposed JacobiNeRF demonstrates coherence between points with similar semantics.}
           \vspace{0.1cm}
           \label{fig:teaser}
        \end{center}
    }]
}

\begin{document}

\title{JacobiNeRF: NeRF Shaping with Mutual Information Gradients
}

\author{
Xiaomeng Xu$^{1,*}$~~~
Yanchao Yang$^{2,3,*,\dag}$~~~
Kaichun Mo$^{3,4}$~~~
Boxiao Pan$^{3}$~~~
Li Yi$^{1,5,6}$~~~
Leonidas Guibas$^{3,7}$
\vspace{0.3cm}
\\
$^1$Tsinghua University~~~
$^2$The University of Hong Kong~~~
$^3$Stanford University~~~
$^4$NVIDIA Research~~~\\
$^5$Shanghai AI Laboratory~~~
$^6$Shanghai Qizhi Institute~~~
$^7$Google Research
\vspace{-0.3cm}
}

\maketitle

\renewcommand{\thefootnote}{\fnsymbol{footnote}}
\footnotetext[1]{Equal Contributions}
\footnotetext[2]{Corresponding Author $<$yanchaoy@hku.hk$>$. The author is also affiliated with the HKU Musketeers Foundation Institute of Data Science.}

%%%%%%%%% ABSTRACT
\begin{abstract}
We propose a method that trains a neural radiance field (NeRF) to encode not only the appearance of the scene but also semantic correlations between scene points, regions, or entities -- aiming to capture their mutual co-variation patterns. 
In contrast to the traditional first-order photometric reconstruction objective, our method explicitly regularizes the learning dynamics to align the Jacobians of highly-correlated entities, which proves to maximize the mutual information between them under random scene perturbations. 
By paying attention to this second-order information, we can shape a NeRF to express semantically meaningful synergies when the network weights are changed by a delta along the gradient of a single entity, region, or even a point. 
To demonstrate the merit of this mutual information modeling, we leverage the coordinated behavior of scene entities that emerges from our shaping to perform label propagation for semantic and instance segmentation. 
Our experiments show that a JacobiNeRF is more efficient in propagating annotations among 2D pixels and 3D points compared to NeRFs without mutual information shaping, especially in extremely sparse label regimes -- thus reducing annotation burden. 
The same machinery can further be used for entity selection or scene modifications. 
Our code is available at \href{https://github.com/xxm19/jacobinerf}{https://github.com/xxm19/jacobinerf}.
\end{abstract}

%%%%%%%%% BODY TEXT
\vspace{-0.4cm}
\section{Introduction}
\label{sec:intro}

When a real-world scene is perturbed, the response is generally local and semantically meaningful, e.g., a slight knock on a chair will result in a small displacement of just that chair. 
Such coherence in the perturbation of a scene evidences high mutual information between certain scene points or entities that can be leveraged to discover instances or semantic groups \cite{yang2019unsupervised,yang2021dystab}. 
A NeRF scene representation, however, solely supervised with 2D photometric loss may not converge to a configuration that reflects the actual scene structure \cite{zhang2020nerf++}; even if the density is correctly estimated, the network in general will {\it not} be aware of the underlying semantic structure.
As shown in Fig.~\ref{fig:teaser},
a perturbation on a specific entity of the scene through the network weights activates almost all other entities. 

This lack of semantic awareness may not be a problem for view synthesis and browsing, but it clearly is of concern when such neural scene representations are employed for interactive tasks that require understanding the underlying scene structure, e.g., entity selection, annotation propagation, scene editing, and so on. All these tasks can be greatly aided by a representation that better reflects the correlations present in the underlying reality.  We take a step towards endowing neural representations with such awareness of the mutual scene inter-dependencies by asking {\it how} it is possible to train a NeRF, so that it {\it not only} reproduces the appearance and geometry of the scene, {\it but also} generates coordinated responses between correlated entities when perturbed in the network parameter space. 

Current approaches that encode semantics largely treat semantic labels (e.g., instance segmentation) \cite{zhi2021place,kundu2022panoptic} as a separate channel, in addition to density or RGB radiance. 
However, in the semantics case, the value of the channel (e.g., instance ID) is typically an artifact of the implementation. 
What really matters is the decomposition of the 2D pixels (or of the scene 3D points) the NeRF encodes into groups -- this is because semantics is more about relationships than values. 
Thus, we introduce an information-theoretic technique whose goal is to ``shape'' an implicit NeRF representation of a scene to better reflect the underlying regularities (``semantics'') of the world;
so as to enforce consistent variation among correlated scene pixels, points, regions, or entities, enabling efficient information propagation within and across views.

The {\it key} to the proposed ``shaping'' technique is an equivalence between mutual information and the normalized inner product (cosine similarity) of the Jacobians at two pixels or 3D points.
More explicitly, if we apply random delta perturbations to the NeRF weights, the induced random values of two pixels share mutual information up to the absolute cosine similarity of their gradients or Jacobians with respect to the weights computed at the unperturbed NeRF.
This theoretical finding ensures a large correlation between scene entities with high mutual information -- and thus coherent perturbation-induced behaviors -- if their tangent spaces are aligned.
Based on this insight, we apply contrastive learning to align the NeRF gradients with general-purpose self-supervised features (e.g., DINO), which is why we term our NeRF \emph{``JacobiNeRF''}. While several prior works \cite{tschernezki22neural,kobayashi2022distilledfeaturefields} distill 1st-order semantic information from 2D views to get a consensus 1st-order feature in 3D, we {\it instead} regularize the NeRF using 2nd-order, mutual information based contrastive shaping on the NeRF gradients to achieve semantic consensus -- now encoded in the NeRF tangent space.

The proposed NeRF shaping sets up resonances between correlated pixels or points and makes the propagation of all kinds of semantic information possible from sparse annotations -- because pixels that co-vary with the annotated one are probably of the same semantics indicated by the mutual information equivalence. 
For example, we can use such resonances to propagate semantic or instance information as shown in Sec.~\ref{sec:label-prop}, where we also show that our contrastive shaping can be applied to gradients of 2D pixels, or of 3D points.
The same machinery also enables many other functions, including the ability to select an entity by clicking at one of its points or the propagation of appearance edits, as illustrated in Fig.~\ref{fig:material_recolor}. 
Additionally, our approach suggests the possibility that a NeRF shaped with rich 2nd-order relational information in the way described may be capable of propagating many additional kinds of semantics without further re-shaping -- because the NeRF coefficients have already captured the essential ``DNA'' of points in the scene, of which different semantic aspects are just different expressions.
In summary, our key contributions are:
\begin{itemize}[leftmargin=*]
\vspace{-1mm}
\item We propose the novel problem of shaping NeRFs to reflect mutual information correlations between scene entities under random scene perturbations.
\vspace{-1mm}
\item We show that the mutual information between any two scene entities is equivalent to the cosine similarity of their gradients with respect to the perturbed weights.
\vspace{-1mm}
\item We develop {\it JacobiNeRF}, a shaping technique that effectively encodes 2nd-order relational information into a NeRF tangent space via contrastive learning.
\vspace{-1mm}
\item We demonstrate the effectiveness of JacobiNeRF with state-of-the-art performance on sparse label propagation for both semantic and instance segmentation tasks.
%\vspace{-1mm}
\end{itemize}

\section{Related Work}
\label{sec:related}

\paragraph{Neural Radiance Fields.}
Recent work has demonstrated promising results in implicitly parametrizing 3D scenes with neural networks. NeRF \cite{mildenhall2021nerf} is one notable work among many others~\cite{zhang2020nerf++,liu2020neural,yu2021pixelnerf,martin2021nerf,park2021nerfies,karras2021alias,muller2022instant,barron2022mip,kangle2021dsnerf,xu2022point} that train deep networks to encode photometric attributes for novel view synthesis. Besides rendering quality, quite a few focus on the reconstructed geometry \cite{wang2021neus,lin2021barf,oechsle2021unisurf,fu2022geo,wei2021nerfingmvs,zhu2023vdnnerf}. There are also attempts towards making NeRF compositional, e.g., GRAF \cite{schwarz2020graf}, CodeNeRF \cite{jang2021codenerf},  CLIP-NeRF \cite{wang2022clip} and PNF \cite{kundu2022panoptic} explicitly model shape and appearance so that one can modify the color or shape of an object by adjusting their separate codes.
EditNeRF \cite{liu2021editing} further extends this direction by exploring different structures and tuning methods for edit propagation via delicate control of sampled rays.
These methods, however, mostly work with object-level or category-level NeRFs in contrast to a whole scene composed of many objects. Moreover, disentanglement in shape and appearance does not necessarily guarantee instance or semantic consistency. For example, \cite{yang2021learning} learns object-compositional neural radiance field for scene editing, yet instance masks are required during training. Our work studies how to further shape a holistic NeRF representation to enable awareness of the underlying semantic structure.

\vspace{-2mm}
\paragraph{Self-supervised Representation Learning.}
Due to the absence of annotations, self-supervised feature learning has attracted much attention in recent literature~\cite{chen2020simple,he2020momentum,grill2020bootstrap,caron2020unsupervised,chen2020big,zbontar2021barlow,he2022masked}. Among these, DINO~\cite{caron2021emerging} proposes to perform self-distillation with a teacher and a student network, demonstrating that the features learned exhibit reasonable semantic information. 
There are also works utilizing self-supervised features for scene partition. For example, \cite{melas2022deep,wang2022self} formulates unsupervised image decomposition as a traditional graph partitioning with affinity matrix computed from self-supervised features. 
Please refer to \cite{khan2022transformers} for a more comprehensive overview of self-supervised feature learning techniques and their applications.
Besides clustering with self-supervised features, 
CIS \cite{yang2019unsupervised,yang2021dystab} directly leverages mutual information learned in an adversarial manner to perform unsupervised object discovery. However, our focus is to encode mutual information in 3D scene representations. Moreover, our work can seamlessly incorporate arbitrary self-supervised features for learning meaningful correlations between scene entities.

\vspace{-2mm}
\paragraph{2D to 3D Feature Distillation \& Label Propagation}
2D information annotated on NeRF 2D views (e.g., semantic labels) can be pushed into the 3D structure of a NeRF \cite{zhi2021place}. And the process can be regularized and improved, for example,
N3F~\cite{tschernezki22neural} treats 2D image features from pretrained networks as additional color channels and train a NeRF to reconstruct the augmented radiance.
DFF~\cite{kobayashi2022distilledfeaturefields} similarly distills knowledge of off-the-shelf 2D image feature extractors into NeRFs with volume rendering. 
In contrast to feature distillation, Semantic-NeRF \cite{zhi2021place} adds a parallel branch along the color one in NeRF \cite{mildenhall2021nerf} to predict semantic logits supervised by annotated pixels or images.
With similar technique, Panoptic-NeRF \cite{fu2022panoptic} encodes coarse and noisy annotations into NeRFs.
Both demonstrate that volume rendering improves 3D consistency, and is capable of denoising and interpolating imperfect 2D labels.
However, as they are first-order, they can only replicate the provided supervision and cannot be used for other tasks.
Note that ADeLA \cite{ren2022adela} also leverages information flow between two frames for label propagation. Since it mainly relies on appearance similarity, the generalization to significant viewpoint change may not be guaranteed.

\section{Method}
\label{sec:method}

We seek scene representations that not only mirror the appearance and geometry of a scene, but also encode mutual correlations between scene regions and entities -- in the sense that the representation facilitates the generation of semantically meaningful scene perturbations which will change such regions and entities in coordinated ways.
Our study here mainly focuses on NeRFs, due to their capability for high-fidelity view synthesis. We show, however, that in the absence of correlation regularization on the learning dynamics, solely reconstructing the scene with standard photometric losses, does {\it not} guarantee semantic synergies between entities under various scene perturbations.

\begin{figure}[!t]
  \centering
  \includegraphics[width=0.95\linewidth]{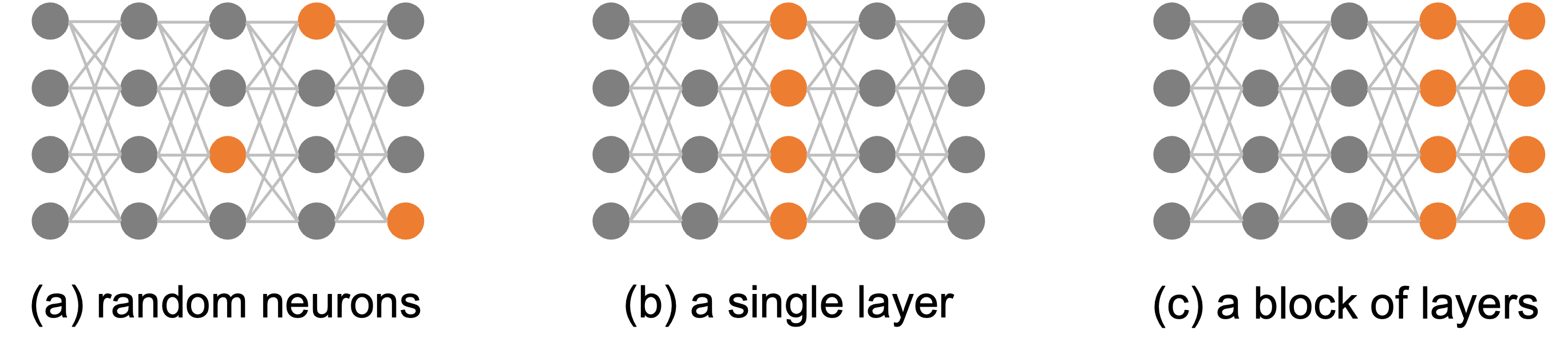}
  \vspace{-0.2cm}
  \caption{We examine mutual information between scene entities induced by perturbing the NeRF weights in different patterns: (a): a set of randomly selected neurons; (b): a single layer of the MLP; (c): a block of layers, e.g., the color branch in the MLP.}
  \vspace{-0.1cm}
  \label{fig:perturbations}
\end{figure}

In order to analyze how NeRFs render mutual correlation under perturbations, 
we derive that \emph{the mutual information (MI) between any two scene entities is equivalent to the cosine similarity of their gradients with respect to the perturbed network parameters}. Based on this equivalence, we propose a training regiment that biases the parameters to properly encode mutual correlation between 3D points or 2D pixels. Our method shapes NeRFs in a way that {\it minimizes} the synthesis discrepancy, while {\it maximizing} the synergy between correlated scene entities when perturbing along the gradients, without changing the architecture. We further show how this mutual-information-induced synergy can be leveraged to propagate labels for both semantic and instance segmentation.

\subsection{NeRF preliminaries}

Given a set of posed images $\{I_k\}$, NeRFs aim at learning an implicit field representation of the scene from which new views can be generated through volume rendering density and radiance values.
Denote $x\in\mathbb{R}^3$ as a point in 3D whose radiance is determined by a color function $c: \mathbb{R}^3\times\mathbb{S}^2\rightarrow\mathbb{R}^3$, mapping the point coordinate $x$ and a viewing direction $\mathbf{v}\in\mathbb{S}^2$ to an RGB value.
Also, denote $\p$ as a pixel on the image plane specified by a camera center $\mathbf{o}$ and a direction $\mathbf{v}$, 
the volume rendering procedure that generates the pixel value of $\p$ can be described by:
\begin{equation}
    I(\p) = \Phi(\mathbf{o},\mathbf{v};\theta)
    = \int\limits_0^{+\infty} w(t;\theta)c(\mathbf{p}(t),\mathbf{v};\theta)\mathrm{d}t\,,
    \label{eq:volume_rend}
\end{equation}
where $\{\mathbf{p}(t)=\mathbf{o}+t\mathbf{v}\mid t\geq 0\}$ is the camera ray passing through the camera center $\mathbf{o}$ and the pixel $\p$,
and $w$ is a weight function computed from the density values. While an immense number of NeRF variations have been explored, for simplicity we use the original and most basic MLP formulation.
Please refer to \cite{mildenhall2021nerf} for more details on volume rendering and $w$, together with how to perform the integration in Eq.~\eqref{eq:volume_rend} under discretization.

The natural way to perturb a scene represented by such a NeRF is to perturb the MLP parameters. Unless otherwise mentioned, all trainable parameters are compacted into the vector $\theta$ in Eq.~\eqref{eq:volume_rend}. Traditional NeRFs losses are 1st-order, optimizing values such as density and color. In this work we focus additionally and crucially on \emph{2nd-order supervision}, optimizing correlations between scene entities when the scene is perturbed through changes in the MLP parameters. In this way we aim to ``shape'' the NeRF MLP to better reflect any supervision we may have on the co-variation structure of the scene contents. 
Next, we detail the perturbations and the derivation of mutual information in an analytical form.

\begin{figure}[!t]
  \centering
  \includegraphics[width=1.0\linewidth]{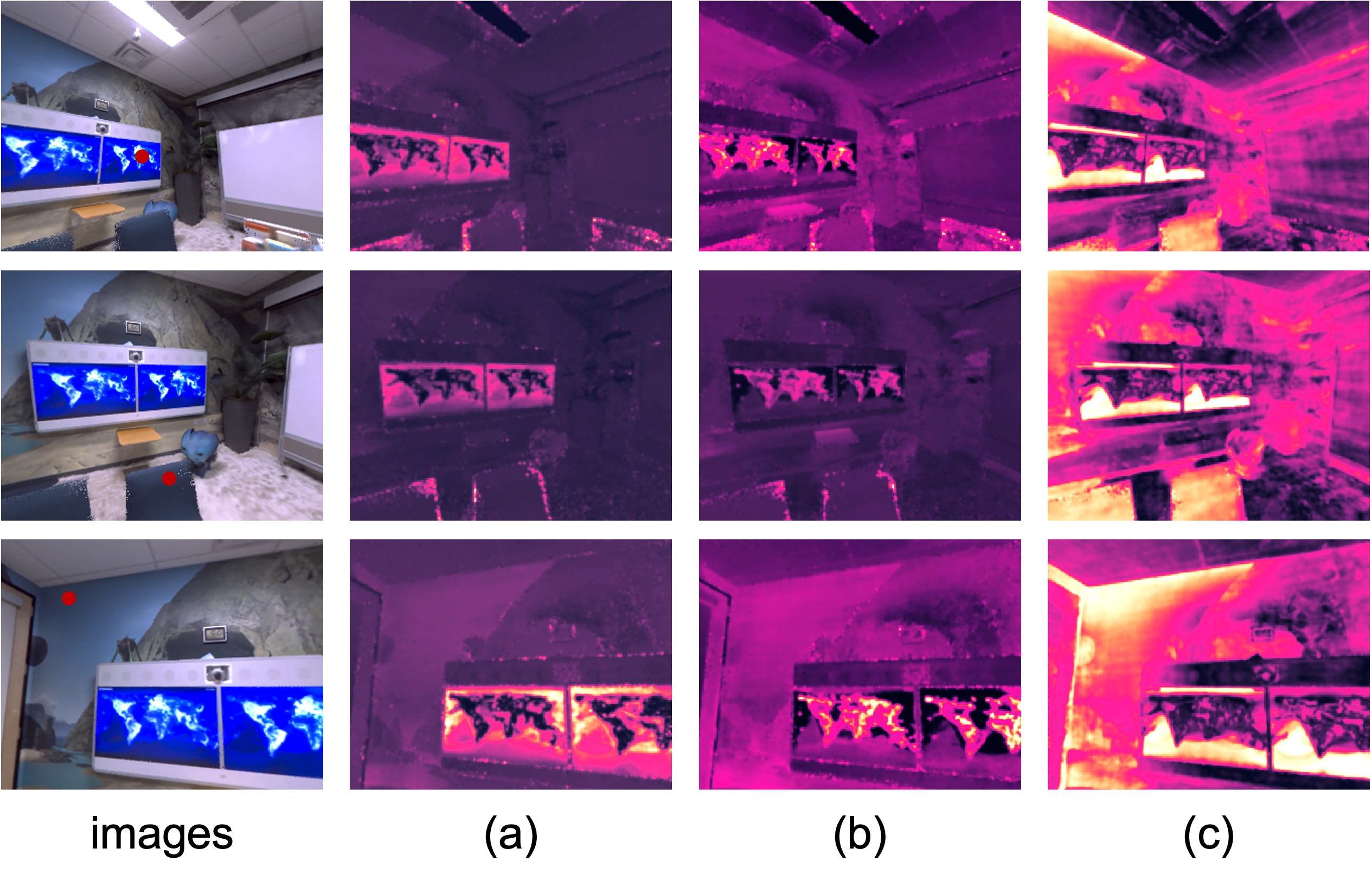}
  \vspace{-0.5cm}
  \caption{Mutual information computed with Eq.~\eqref{eq:mutual-info} under different perturbation patterns. (a): a set of randomly selected neurons; (b): a single MLP layer; (c): a block of layers, e.g., the color branch. As observed, none of the perturbation patterns induces correlations that are localized and semantically meaningful.}
  \vspace{-0.2cm}
  \label{fig:MI-different-perturbs}
\end{figure}

\subsection{Mutual information approximation by Jacobian inner products}

We study the correlation under MLP parameter perturbations of the values produced by the NeRF, either at 3D points, or (after volume rendering) at 2D view pixels. For concreteness, we focus on two pixels $\p_i$ and $\p_j$ and their gray-scale values $I(\p_i)$ and $I(\p_j)$, respectively (the pixels may or may not come from the same view): 
\begin{align*}
    I(\p_i) &= \Phi(\co_i,\cv_i;\theta)\,,\\
    I(\p_j) &= \Phi(\co_j,\cv_j;\theta)\,.
\end{align*}

Further, we denote by $\theta^D$  the set of parameters that will be perturbed by a random noise vector $\mathbf{n}\in\mathbb{R}^D$ sampled from a uniform distribution on the sphere $\mathbb{S}^{D-1}$. Please see Fig.~\ref{fig:perturbations} for different selection patterns of $\theta^D$.
The random variables representing the perturbed pixel values are then:
\begin{align*}
    \hat{I}(\p_i) &= \Phi(\co_i,\cv_i;\theta^D+\n)\,,\\
    \hat{I}(\p_j) &= \Phi(\co_j,\cv_j;\theta^D+\n)\,,
    \label{eq:randV}
\end{align*}
and we omit parameters that remain unchanged for clarity.

\begin{figure*}[!t]
  \centering
  \includegraphics[width=0.96\linewidth]{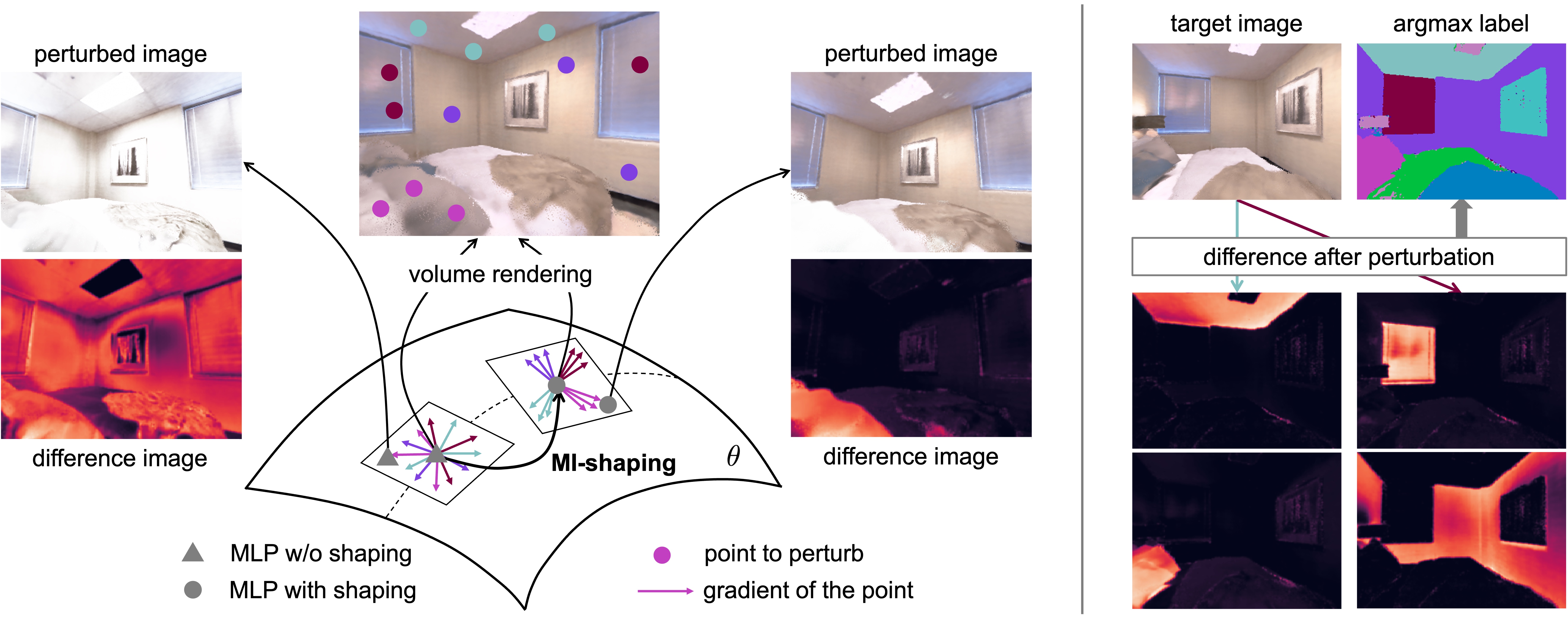}
\vspace{-0.2cm}
  \caption{{\it Left}: Mutual information shaping of a NeRF in the tangent space. The pre-shaping NeRF (triangle on the manifold) has a tangent space where the Jacobians of semantically similar points distribute randomly, so that a perturbation along the gradient of a point from the pillow induces changes all over the scene. After MI-shaping, the NeRF (circle on the manifold) can still render the same scene, but now the Jacobians in the tangent space are consistently distributed. Thus the same perturbation only affects the scene region corresponding to the selected pillow. {\it Right}: Given a target (unlabeled) view of the scene, we can generate labels for it by taking the argmax of the perturbation responses from those annotated in the source view (dots in the image above the manifold) leveraging the post-shaping resonances.}
  \vspace{-0.4cm}
  \label{fig:training-propagate}
\end{figure*}

Now we characterize the mutual information between $\hat{I}(\p_i)$ and $\hat{I}(\p_j)$ under the perturbation-induced joint probability distribution $\mathbb{P}(\hat{I}(\p_i),\hat{I}(\p_j))$.
However, calculating the joint distribution under a push-forward of the MLP is complicated due to non-linearities.
Thus, we proceed by constraining the magnitude of the perturbations, namely, multiplying the random noise $\n$ by $\sigma\ll 1.0$.
This constraint improves compliance with the fact that a small perturbation in the physical scene is enough to reveal mutual information between scene entities.
For example, slightly pushing a chair can generate a motion that shows correlations between different parts of the environment. 
Moreover, it makes it likely that the perturbed representation still represents a legitimate scene.
With this constraint,
we can explicitly write the random variables under consideration as:
\begin{align}
    \hat{I}(\p_i) &= I(\p_i) + \sigma\n\cdot \dfrac{\partial \Phi(\co_i,\cv_i;\theta)}{\partial \theta^D}\,,\\
    \hat{I}(\p_j) &= I(\p_j) + \sigma\n\cdot \dfrac{\partial \Phi(\co_j,\cv_j;\theta)}{\partial \theta^D}\,,
\end{align}
following a Taylor expansion. 
We denote the respective Jacobians as $\partial \Phi_i/\partial \theta^D$ or $\partial \Phi_i$ for notational ease.
We can then show that the mutual information is:
\begin{multline}
    \mi(\hat{I}(\p_i),\hat{I}(\p_j)) = \ent(\hat{I}(\p_j)) - \ent(\hat{I}(\p_j)\mid \hat{I}(\p_i))\\
    = \ent(\sigma\n\cdot\partial\Phi_j) - \ent(\sigma\n\cdot\partial\Phi_j \mid \sigma\n\cdot\partial\Phi_i)\,,
    \label{eq:mi}
\end{multline}
leveraging the fact that entropy is translation-invariant.
Furthermore, by writing the random noise and the Jacobians in spherical coordinates, we can derive that: 
\begin{equation}
    \mi(\hat{I}(\p_i),\hat{I}(\p_j)) = \log(\dfrac{1}{\sqrt{1-\cos^2\alpha}}) + \mathrm{const.}
    \label{eq:mutual-info}
\end{equation}
Here, $\alpha$ is the angle between $\partial\Phi_i$ and $\partial\Phi_j$.
For more derivation details, please refer to the Appendix.
The {\it key} insight is that the {\it mutual information} between the perturbed pixels is {\it positively correlated} with the absolute value of the cosine similarity of their gradients with respect to the perturbed parameters.
More explicitly,
if $\partial\Phi_i$ and $\partial\Phi_j$ are pointing at the same or opposite direction, i.e., $\|\cos\alpha\|$ is close to 1, then the mutual information between $\hat{I}(\p_i)$ and $\hat{I}(\p_j)$ becomes infinity (maximized).
Otherwise, if $\partial\Phi_i$ and $\partial\Phi_j$ are perpendicular, i.e., $\|\cos\alpha\|$ is 0, then the mutual information between the perturbed pixels is minimized.

With this analytical expression of the mutual information, we can efficiently check how different entities from the same scene are correlated.
As observed in Fig.~\ref{fig:MI-different-perturbs},
NeRF weights obtained solely by the reconstruction loss do not reveal meaningful correspondences.
Next, we describe our method that biases the training dynamics of NeRFs so that the shaped weights not only reconstruct the scene well but also reflect the mutual correlation between scene entities.

\subsection{Shaping neural radiance fields with mutual \\information gradients}
\label{sec:mig-shaping}

According to Eq.~\eqref{eq:mutual-info}, 
if we want to render two pixels or points correlated (with high mutual information), 
we can align their gradients regarding the perturbed parameters.
On the other, for entities that share little mutual information, we like their gradients to be orthogonal.
Suppose $(\p_i,\p_{i^+})$ is a pair of highly correlated pixels, whereas $(\p_i,\p_{i^-})$ are independent, then we should observe:
\begin{equation}
    \dfrac{\|\partial\Phi_i^T\partial\Phi_{i^+}\|}{\|\partial\Phi_i\|\|\partial\Phi_{i^+}\|} > \dfrac{\|\partial\Phi_i^T\partial\Phi_{i^-}\|}{\|\partial\Phi_i\|\|\partial\Phi_{i^-}\|}\,.
\end{equation}

Since our goal is to encode the relative correlations between different pairs of scene entities instead of the exact mutual information values, it suffices to minimize the InfoNCE \cite{oord2018representation} loss with positive and negative gradient pairs:
\begin{multline}
    \loss_{\mathrm{MIG}} = -\log\dfrac{\exp(\|\cos(\partial\Phi_i,\partial\Phi_{i^+})\|/\tau)}{\sum_{i^+\cup\{i^-\}}\exp(\|\cos(\partial\Phi_i,\partial\Phi_{i^-})\|/\tau)}\,,
\end{multline}
where $\tau$ is the temperature and we (ab)use $\cos$ for cosine similarity. Note, $\loss_{\mathrm{mig}}$ encourages highly correlated (positive) points to have large cosine similarity by the pull through the numerator.

The question now is how to select positive and negative gradient pairs -- which seems require knowledge about the mutual information between scene entities.
Unfortunately,
since the posed images for training NeRFs are from a (static) snapshot of the physical scene, the joint distribution needed to compute the mutual information is difficult to recover.
This necessitates that we resort to external sources for surrogates of the mutual information, for which, fortunately, we have several candidates.
For example, self-supervised features can come from contrastive learning methods.
We choose DINO features \cite{caron2021emerging} as our primary surrogate due to their capability to capture semantic similarity while maintaining reasonable discriminability. We can of course also accept direct supervision through off-the-shelf external semantic and instance segmentation tools. An interesting issue that we investigate is exactly how much such supervision is needed for meaningful NeRF shaping. We detail the selection process and our results in Sec.~\ref{sec:experiments}.

With positive and negative samples, 
we can write the training loss that endows NeRFs with mutual-information-awareness in the tangent space (perturbations) as:
\begin{equation}
    \loss_{\mathrm{TM}} = \loss_{\mathrm{NeRF}} + \lambda\loss_{\mathrm{MIG}} + \gamma(1.0-\|\partial\Phi_i\|)^2\,,
    \label{eq:shaping-loss}
\end{equation}
where $\loss_{\mathrm{NeRF}}$ is for photometric reconstruction, and $\loss_{\mathrm{MIG}}$ shapes NeRF with mutual information (MI-shaping) through cosine similarity of gradients. 
The last term improves the training efficiency by compacting the gradients onto a unit sphere, which also facilitates label propagation with perturbation response in the following. 
Please see Fig.~\eqref{fig:training-propagate} (left) for a visual illustration of MI-shaping.

We term our 2nd-order NeRF \emph{``JacobiNeRF''} exactly because of the use of the inner products of Jacobians to define these 2nd-order losses that capture mutual information. Our MI-shaping regiment can be thought of a NeRF operator, acting to better align the tangent space of a given NeRF with information we have on mutual correlations between scene pixels, points, regions, or entities.

\subsection{Label propagation with JacobiNeRF}
\label{sec:label-prop}

Besides revealing correlations under small perturbations,
we can also leverage the synergy between different scene entities in a JacobiNeRF to perform label propagation 
by either transporting annotations from one view to another or to densify labels from a few annotated points to the remaining ones.
Next, we illustrate the propagation procedure from a source view $I(\p^s_i) = \Phi^\jb(\mathbf{o}^s,\mathbf{v}^s_i;\theta)$ to a target $I(\p^t_i) = \Phi^\jb(\mathbf{o}^t,\mathbf{v}^t_i;\theta)$ with semantic segmentation -- note that the same method can be directly applied to instance segmentation as well as to different combinations of views.

Suppose we have a list of labeled pixels from the source view $\{(\p^s_k,l^s_k)\}_{k=1...K}$ with one label for each of the K classes, i.e., $l^s_k=k$.
The goal is to determine the semantic labels for every pixel $\p^t_i$ of the target view.
In principle, we can perform a maximum a posteriori (MAP) estimation for each target pixel by:
\begin{equation}
    l^t_i = \argmax_{\hat{l}^t_i} \mathbb{P}(\hat{l}^t_i\mid \{(\p^s_k,l^s_k)\}, \Phi^\jb)\,.
    \label{eq:prop-map}
\end{equation}
If the mutual-information shaping described in Sec.~\ref{sec:mig-shaping} converges properly, we can assume conditional independence between uncorrelated entities. Then the target label $l^t_i$ depends only on the source pixel $\p^s_{k^*}$ which conveys the maximum mutual information towards $\p^t_i$.
Thus, it is legitimate to assign $l^s_{k^*}$ or $k^*$ to $\p^t_i$ in order to maximize the posterior in Eq.~\eqref{eq:prop-map}.
In other words, $l^t_i = l^s_{k^*}$, so that:
\begin{align}
    k^* &= \argmax_k \mi( \Phi^\jb(\mathbf{o}^s,\mathbf{v}^s_k;\theta),
    \Phi^\jb(\mathbf{o}^t,\mathbf{v}^t_i;\theta))\,,\\
    &= \argmax_k \dfrac{\|\partial\Phi^{\jb,s}_k\cdot\partial\Phi^{\jb,t}_i\|}{\|\partial\Phi^{\jb,s}_k\|\|\partial\Phi^{\jb,t}_i\|}\,.
    \label{eq:perturb-MI}
\end{align}

As encouraged by the third term of the shaping loss in Eq.~\eqref{eq:shaping-loss}, the norm of the gradients should be close to $1.0$, which allows us to approximate the cosine similarity between $\partial\Phi^{\jb,s}_k$ and all $\partial\Phi^{\jb,t}_i$'s by a single delta perturbation along $\partial\Phi^{\jb,s}_k$ as evidenced by the Taylor expansion.
Namely,
for each of the labeled pixels, we first generate a perturbed JacobiNeRF along its gradient, i.e.,
\begin{equation}
    \Phi^{\jb}(:;\theta+\sigma\partial\Phi^{\jb,s}_k), k=1...K\,.
    \label{eq:perturbed_term}
\end{equation}
Then each of the perturbed JacobiNeRFs will be used to synthesize a perturbed image in the target view:
\begin{equation}
    I_k(\p^t_i) = \Phi^{\jb}(\co^t,\cv^t_{\p^t_i};\theta+\sigma\partial\Phi^{\jb,s}_k), k=1...K\,.
\end{equation}
Next, we calculate the perturbation response as the absolute difference between the perturbed and original target images:
\begin{equation}
    R_k(\p^t_i) = \ \mid I_k(\p^t_i) - I(\p^t_i)\mid, k=1...K\,.
    \label{eq:difflogits}
\end{equation}
Finally, since the perturbation response $R_k(\p^t_i)$ resembles the mutual information between $I(\p^s_k)$ and $I(\p^t_i)$ (Eq.~\eqref{eq:perturb-MI}), 
we can treat the concatenation $[R_k(\p^t_i)]$ as the logits for a K-way semantic segmentation on the target image.
Thus, we obtain the semantic label for $\p^t_i$ as: $l^t_i = \argmax_{k} R_k(\p^t_i)$.

Please note that the propagation principle discussed above is applicable to any task that is view-invariant. 
For example, K can be the number of entity instances in the scene, and we perform propagation for instance segmentation.
Here, the absolute difference logits are computed in the 2D domain (after volume rendering) in Eq.~\eqref{eq:difflogits}.
However, we can also measure the perturbation differences in 3D first, e.g., 
estimate difference logits for a 3D point along the ray emanating from a certain pixel (following the hierarchical sampling strategy of \cite{mildenhall2021nerf}),
and then leverage volume rendering to accumulate the sampled perturbation differences and arrive at a 2D logit.
In this 3D case, noise in the perturbation may be averaged out, improving the result. We name the results obtained directly in 2D as J-NeRF 2D while the latter as J-NeRF 3D.

\section{Experiments}
\label{sec:exp}
We first describe the datasets in Sec.~\ref{subsec:datasets}, which we use for evaluating JacobiNeRF on semantic and instance label propagation in Sec.~\ref{subsec:exp-semseg} and Sec.~\ref{exp-instseg}, respectively. 
We then perform an extensive ablation in Sec.~\ref{subsec:ablation} on the hyper-parameters. 
Results show that JacobiNeRF is effective in propagating annotations with the encoded correlations, especially in the very sparse label regimes.

\subsection{Datasets}
\label{subsec:datasets}
\vspace{-1mm}
\noindent {\bf Replica} \cite{straub2019replica} is a synthetic indoor dataset with high-quality geometry, texture, and semantic annotations.
We pick the 7 scenes selected by \cite{zhi2021place} for fair comparison.
Each of the scenes comes with 900 posed frames, which are partitioned into a training set and a test set of 180 frames respectively.

\noindent {\bf ScanNet} \cite{dai2017scannet} is a real-world RGB-D indoor dataset. Again, we uniformly sample training and test frames, and ensure that the training and test sets do not overlap and contain approximately the same number of frames (from 180 to 200).

\begin{figure*}[!t]
  \centering
  \includegraphics[width=.9\linewidth]{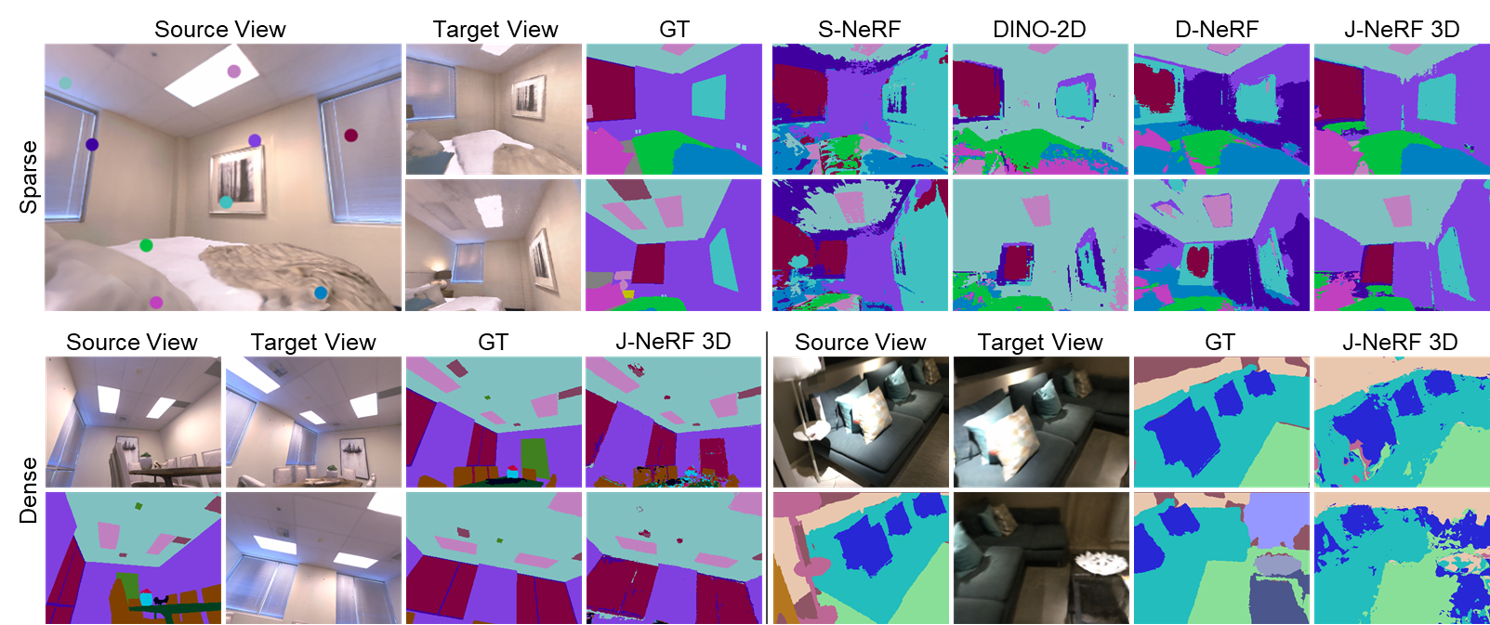}
%  \vspace{-0.6cm}
  \caption{Qualitative results of label propagation for semantic segmentation. Top: propagating sparse labels (colored dots) to different views of the same scene. Bottom: one-view dense label propagation on Replica (left) and ScanNet (right).}
  \label{fig:sem_vis}
  \vspace{-0.3cm}
\end{figure*}

\subsection{Label propagation for semantic segmentation}
\label{subsec:exp-semseg}

\noindent {\bf Experimental setting.}
The training of JacobiNeRF follows Sec.~\ref{sec:mig-shaping}, and for test-time label propagation, we implement two schemes, i.e., J-NeRF 2D and J-NeRF 3D following Sec.~\ref{sec:label-prop}. Please also refer to the appendix for more details.

\label{sec:experiments}
We test the effectiveness of segmentation label propagation under two settings. 
In the \textit{sparse} setting, we only provide a single randomly selected pixel label for each class in the source view, simulating the real-world annotation scenario of user clicking. While for the \textit{dense} setting, all pixels in the source view are annotated. This is useful when we want to obtain fine-grained labels with a higher annotation cost. The propagation strategy for the sparse setting is elaborated in Sec.~\ref{sec:label-prop}. For the dense setting, we employ an adaptive sampling strategy to select the most representative gradients for each class to save possibly redundant perturbations. Furthermore, to best leverage the given dense labels, we apply a lightweight decoding MLP to enhance the argmax operator described in Sec.~\ref{sec:label-prop}.
More details can be found in the appendix.

We compare JacobiNeRF (\textit{J-NeRF}) to Semantic-NeRF \cite{zhi2021place} (\textit{S-NeRF}), which adds a semantic branch to the original NeRF, and hence predicts semantic labels from the color feature integrated into the radiance field. We also compare to \textit{DINO-2D}, which extracts DINO features from the images and propagates with DINO feature similarity; and DINO-NeRF \cite{kobayashi2022distilledfeaturefields} (\textit{D-NeRF}), which distills DINO features to a DINO branch appended to NeRF, and propagates labels with feature similarity of the volume rendered DINO features.
All methods are tested given the same source view labels.
We evaluate with three standard metrics. Namely, mean intersection-over-union (mIoU),  averaged class accuracy, and total accuracy. The scores are obtained by averaging all test views for each scene. Since we only provide sparse or dense labels from one view, some classes in the test views may not be seen from the source view. Therefore, we exclude them and only evaluate with the seen classes.

\begin{figure}[!t]
  \centering
  \includegraphics[width=1.0\linewidth]{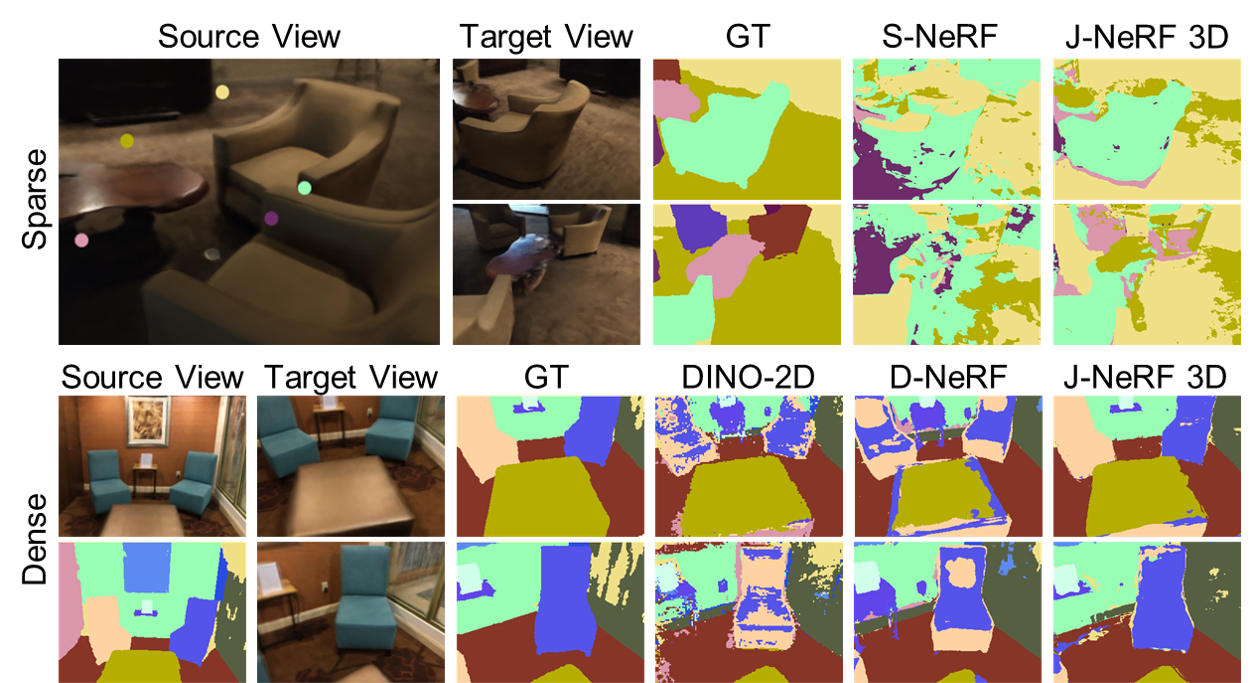}
  \vspace{-0.6cm}
  \caption{Qualitative results of instance segmentation propagation. Examples are from the ScanNet dataset.}
  \vspace{-0.2cm}
  \label{fig:ins_vis}
\end{figure}

\begin{table}
\centering
\footnotesize
\setlength\tabcolsep{2pt}
\begin{tabular}{ccccccc}
    \toprule
    \multicolumn{2}{c}{Method} & S-NeRF & DINO-2D & D-NeRF & J-NeRF 2D & J-NeRF 3D\\
    \midrule
    S. & mIoU $\uparrow$ & 0.187 & 0.181 & 0.253 & \underline{0.263} & \textbf{0.283} \\
    & Avg Acc $\uparrow$ & 0.461 & 0.461 & \textbf{0.527} & 0.489 & \underline{0.524} \\
    & Total Acc $\uparrow$ & 0.310 & 0.414 & 0.407 & \underline{0.483} & \textbf{0.503} \\
    \midrule
    D. & mIoU $\uparrow$ & \underline{0.523} & 0.335 & 0.403 & 0.446 & \textbf{0.524} \\
    & Avg Acc $\uparrow$ & \textbf{0.728} & 0.624 & 0.654 & 0.619 & \underline{0.689} \\
    & Total Acc $\uparrow$ & \underline{0.766} & 0.714 & 0.683 & 0.751 & \textbf{0.864} \\
    \bottomrule
\end{tabular}
\vspace{-0.2cm}
\caption{Semantic segmentation propagation on Replica in sparse (S) and dense (D) settings.}
\label{tab:sem_result}
\vspace{-0.4cm}
\end{table}

\noindent {\bf Results.} 
Tab.~\ref{tab:sem_result} summarizes the quantitative results for semantic segmentation propagation on the 7 scenes from the Replica \cite{straub2019replica} dataset. Our method consistently achieves the best performance in both sparse and dense settings by utilizing correlations encoded in the tangent space.

Fig.~\ref{fig:sem_vis} compares JacobiNeRF with the baselines qualitatively. As observed, with sparse annotations, JacobiNeRF propagates to novel views with much better quality and smoothness than the baselines. In the dense setting, JacobiNeRF demonstrates the ability to propagate labels at a finer granularity on both synthetic and real-world data.

\subsection{Label propagation for instance segmentation}
\label{exp-instseg}

\noindent {\bf Experimental setting.} 
The training and test settings are the same as the semantic segmentation procedure in Sec.~\ref{subsec:exp-semseg}.

\begin{table}
\centering
\footnotesize
\setlength\tabcolsep{2pt}
\begin{tabular}{ccccccc}
    \toprule
    \multicolumn{2}{c}{Method} & S-NeRF & DINO-2D & D-NeRF & J-NeRF 2D & J-NeRF 3D\\
    \midrule
    S. & mIoU $\uparrow$ & 0.154 & 0.206 & 0.191 & \underline{0.299} & \textbf{0.332}\\
    & Avg Acc $\uparrow$ & 0.313 & 0.355 & 0.357 & \underline{0.486} & \textbf{0.525}\\
    & Total Acc $\uparrow$ & 0.327 & 0.362 & 0.372 & \underline{0.519} & \textbf{0.547}\\
    \midrule
    D. & mIoU $\uparrow$ & \textbf{0.421} & 0.344 & 0.353 & 0.400 & \textbf{0.421} \\
    & Avg Acc $\uparrow$ & \textbf{0.619} & 0.525 & 0.541 & 0.535 & \underline{0.558}\\
    & Total Acc $\uparrow$ & 0.603 & 0.625 & 0.620 & \underline{0.644} & \textbf{0.671}\\
    \bottomrule
\end{tabular}
\vspace{-0.2cm}
\caption{Label propagation for instance segmentation on ScanNet with sparse (S.) and dense (D.) annotations.}
\label{tab:in_result}
\vspace{-0.2cm}
\end{table}

\begin{table}
%\vspace{2mm}
\centering
\footnotesize
\setlength\tabcolsep{1.4pt}
    \begin{tabular}{cccc}
    \cline{1-4} 
    \toprule
    \multicolumn{2}{c}{View Distance} & Close & Far\\
    %\cline{1-4} 
    \midrule
    S-NeRF & mIoU $\uparrow$ & 0.77 & 0.28\\
    & Total Acc $\uparrow$ & 0.91 & 0.58\\
    %\cline{1-4} 
    %\specialrule{0em}{0.5pt}{0pt}
    \midrule
    J-NeRF & mIoU $\uparrow$ & 0.75 & 0.47\\
    & Total Acc $\uparrow$ & 0.93 & 0.83\\
    %\cline{1-4} 
    %\specialrule{0em}{0.5pt}{0pt}
    \bottomrule
    \end{tabular}
    \quad
    \begin{tabular}{ccccc}
    %\cline{1-5} 
    %\specialrule{0em}{0.5pt}{0pt}
    \toprule
    \multicolumn{2}{c}{View Number} & 1 & 2 & 3\\
    %\cline{1-5} 
    %\specialrule{0em}{0.5pt}{0pt}
    \midrule
    S-NeRF & mIoU $\uparrow$ & 0.18 & 0.38 & 0.57\\
    & Total Acc $\uparrow$ & 0.64 & 0.85 & 0.90\\
    %\cline{1-5} 
    %\specialrule{0em}{0.5pt}{0pt}
    \midrule
    J-NeRF & mIoU $\uparrow$ & 0.22 & 0.38 & 0.55\\
    & Total Acc $\uparrow$ & 0.75 & 0.90 & 0.92\\
    %\cline{1-5} 
    %\specialrule{0em}{0.5pt}{0pt}
    \bottomrule
    \end{tabular}
    \vspace{-0.2cm}
    \caption{Left: propagation performance on near and distant views (dense).
    Right: comparison with multiview dense supervision.
    }
\vspace{-0.5cm}
\label{tab:far_multi}
\end{table}

\noindent {\bf Results.} In Tab.~\ref{tab:in_result}, we show the results of instance segmentation propagation on 4 scenes from the ScanNet dataset. We compare with the same baselines as in the semantic segmentation task. In this more challenging setting again JacobiNeRF significantly outperforms the other baselines on all metrics in the sparse setting, and is comparable with Semantic-NeRF \cite{zhi2021place} in the dense setting  -- though across the board performance is lower than in the semantic segmentation case.
Fig.~\ref{fig:ins_vis} shows the qualitative results under both sparse and dense settings. Our scheme demonstrates the capability to discriminate between different instances of the same class and correctly transport given labels.

\vspace{-0.1cm}
\subsection{Ablation study}
\label{subsec:ablation}

\begin{wrapfigure}{l}{4.5cm}
\centering
  \includegraphics[width=1.0\linewidth]{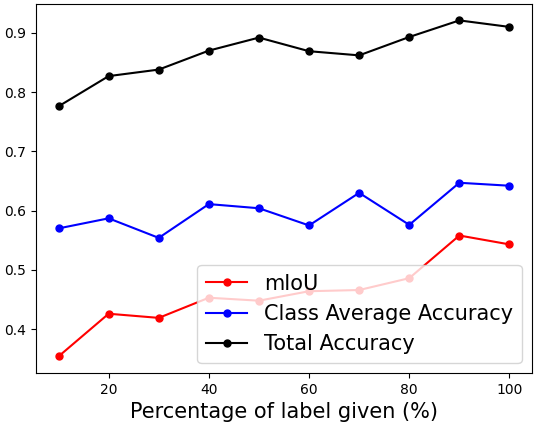}
  \vspace{-0.7cm}
  \caption{Propagation performance under different label densities.}
  \vspace{-0.15cm}
  \label{fig:density-curve}
\end{wrapfigure}
\noindent {\bf Label density.}
Fig.~\ref{fig:density-curve} shows the semantic label propagation performance of our method on one scene under various label density settings. We provide dense labels from one view, and randomly select a sub-region of labels for each class, following the setting in \cite{zhi2021place} with varying density. The horizontal axis denotes the percentage (in area) of used labels for each class. As the density of the label increases, the propagation performance also gets better.

\noindent {\bf Views far from the source view.}
We find that S-NeRF overfits to the source view with low-quality propagation on distant views. 
In contrast, J-NeRF generalizes much better as the shaping can be applied on all views (Tab.~\ref{tab:far_multi}, left).

\noindent {\bf Multiview dense supervision.}
We report the results with multiview supervision in Tab.~\ref{tab:far_multi} (right). 
The gap between S-NeRF and J-NeRF decreases as the view number increases, 
due to the sub-optimal downsampling of dense labels with limited computation. 
A more sufficient and efficient downsampling scheme is our next goal.

\begin{wrapfigure}{l}{4.5cm}
  \centering
  \includegraphics[width=\linewidth]{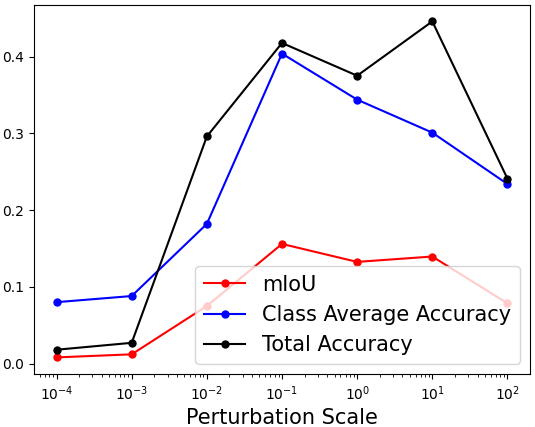}
  \vspace{-0.7cm}
  \caption{Effect of the perturbation magnitude $\sigma$ for label propagation.}
  \label{fig:perturbation-scale}
\end{wrapfigure}
\vspace{2mm}
\noindent {\bf Perturbation magnitude.}
We study how different perturbation scales, i.e., the $\sigma$ in Eq.~\ref{eq:perturbed_term}, affect the label propagation performance. Fig.~\ref{fig:perturbation-scale} shows the results of J-NeRF 2D in the sparse setting. As observed, if the magnitude is too small, the propagation is not good due to weak (noisy) responses. However, if the magnitude is too large, the approximation with Taylor expansion in Eq.~\ref{eq:randV} becomes invalid, thus producing worse results. So we treat it as a hyper-parameter and empirically set $\sigma=0.1$ for all runs.

\vspace{-0.05cm}
\begin{wrapfigure}{r}{4.6cm}
    \includegraphics[width=\linewidth]{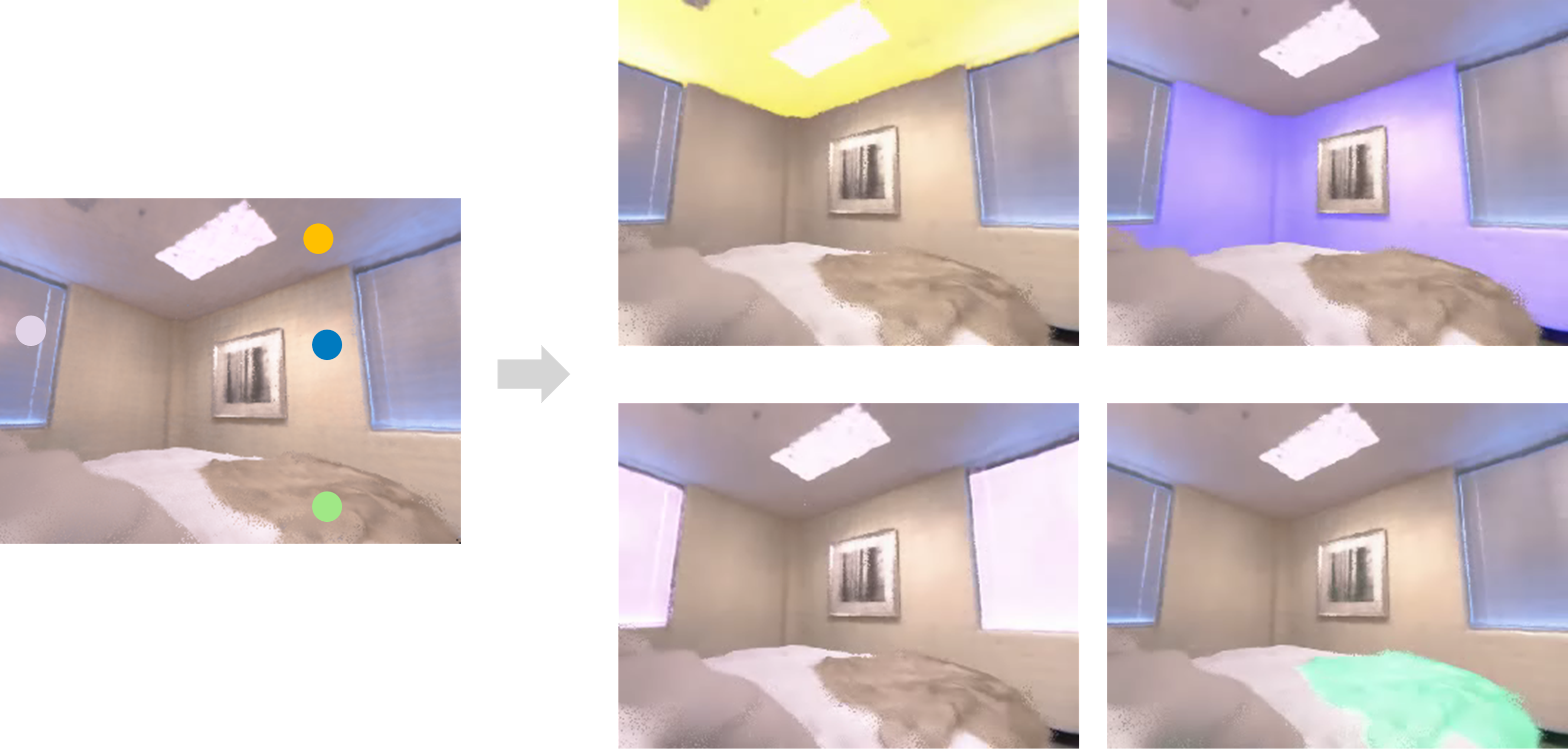}
    \vspace{-0.7cm}
    \caption{Scene re-coloring.}
    \vspace{-0.15cm}
\label{fig:material_recolor}
\end{wrapfigure}
\subsection{Beyond label propagation} Our approach can also be used to propagate or edit other kinds of information. Remarkably, the emerged resonances from MI-shaping allow re-coloring an entire semantic entity (Fig.~\ref{fig:material_recolor}) by perturbing just one of its pixels along the RGB channels.

\vspace{-0.1cm}
\section{Conclusion}

We have demonstrated a way to regularize the learning dynamics of a NeRF, so it reflects the correlations between high mutual information entities. 
This is achieved through the MI-shaping technique that aligns gradients of these correlated entities with respect to the network parameters via a contrastive learning. 
Once shaped in the tangent space with second-order training, the induced resonances in
a NeRF can propagate value information from selected quantities to other correlated quantities leveraging the mutual-information-induced synergies. 
We demonstrate this capability by the propagation of semantic and instance information, as well as semantic entity selection and editing. 
Currently, the shaping scheme relies on self-supervised visual features, but it can be easily oriented to consume features from foundation models to encode cross-modal mutual correlations.
We expect more correlation-informed applications to be possible using the proposed mutual information gradient alignment techniques for NeRFs.
% neural scene representations.

\vspace{0.3cm}
\noindent {\bf Acknowledgement}: We thank the support of a grant from the Stanford Human-Centered AI Institute, an HKU-100 research award, and a Vannevar Bush Faculty Fellowship.

\section{Appendix}

\subsection{Mutual information approximated by Jacobian under scene perturbations}

As defined in Sec.~\ref{sec:method},
$\p_i$ and $\p_j$ are different pixels with values $I(\p_i)$ and $I(\p_j)$, respectively,
and they may or may not come from the same image (view).
Generally speaking, $\p_i$ and $\p_j$ can also be the radiance value of 3D points. Without loss of generality, we derive with (gray-scale) 2D pixels in the following, and they can be expressed as:
\begin{align*}
    I(\p_i) &= \Phi(\co_i,\cv_i;\theta)\\
    I(\p_j) &= \Phi(\co_j,\cv_j;\theta)
\end{align*}

Further, we denote $\theta^D$ as the set of parameters that will be perturbed by a random noise $\mathbf{n}\in\mathbb{R}^D$ sampled from a uniform distribution on the sphere $\mathbb{S}^{D-1}$.
Please see Fig.~\ref{fig:perturbations} in the main text for different selection patterns of $\theta^D$.
The random variables representing the perturbed pixel values are then:
\begin{align*}
    \hat{I}(\p_i) &= \Phi(\co_i,\cv_i;\theta^D+\n)\\
    \hat{I}(\p_j) &= \Phi(\co_j,\cv_j;\theta^D+\n)
    \label{eq:randV}
\end{align*}
where we omit the parameters remain unchanged for clarity.

Now we characterize the mutual information between $\hat{I}(\p_i)$ and $\hat{I}(\p_j)$ under the perturbation-induced joint probability distribution $\mathbb{P}(\hat{I}(\p_i),\hat{I}(\p_j))$.
However, as mentioned, calculating the joint distribution under a push-forward of the MLP is complicated due to non-linearities.
Thus, we proceed by constraining the magnitude of the perturbations, namely, multiplying the random noise $\n$ by $\sigma\ll 1.0$.
Again, this constraint improves compliance with the fact that a small perturbation in the physical scene is enough to reveal mutual information between scene entities.
Moreover, it guarantees that the perturbed representation still represents a legitimate scene.
With this constraint,
we can explicitly write the random variables under consideration as:
\begin{align*}
    \hat{I}(\p_i) &= I(\p_i) + \sigma\n\cdot \dfrac{\partial \Phi(\co_i,\cv_i;\theta)}{\partial \theta^D}\\
    \hat{I}(\p_j) &= I(\p_j) + \sigma\n\cdot \dfrac{\partial \Phi(\co_j,\cv_j;\theta)}{\partial \theta^D}
\end{align*}
following the Taylor expansion. 
We denote the Jacobians as $\partial \Phi_i/\partial \theta^D$ or $\partial \Phi_i$ for easy notation.
We can show that the mutual information is:
\begin{multline*}
    \mi(\hat{I}(\p_i),\hat{I}(\p_j)) = \ent(\hat{I}(\p_j)) - \ent(\hat{I}(\p_j)\mid \hat{I}(\p_i))\\
    = \ent(\sigma\n\cdot\partial\Phi_j) - \ent(\sigma\n\cdot\partial\Phi_j \mid \sigma\n\cdot\partial\Phi_i)
    %\label{eq:mi}
\end{multline*}
leveraging the fact that entropy is translation-invariant.

From the above equation, we can see that the mutual information between the two perturbed pixels can be approximated by the mutual information between two random projections, e.g., $\sigma\n\cdot\partial\Phi_j$ and $\sigma\n\cdot\partial\Phi_i$. To further ease annotation, we let $\A=\partial \Phi_i$ and $\B=\partial \Phi_j$. And,
\begin{equation*}
    \mi(\hat{I}(\p_i),\hat{I}(\p_j)) = \ent(\sigma\n\cdot\B) - \ent(\sigma\n\cdot\B \mid \sigma\n\cdot\A)
\end{equation*}

To proceed, we write the Jacobians and the random (small) perturbation in the spherical form:
\begin{multline}
    \A=\sigma_A\begin{pmatrix}1 \\0 \\\vdots \\0\end{pmatrix}, \B=\sigma_B\begin{pmatrix}\cos\alpha_1^B \\\sin\alpha_1^B\cos\alpha_2^B \\\sin\alpha_1^B\sin\alpha_2^B\cos\alpha_3^B \\\vdots \\\sin\alpha_1^B...\sin\alpha_{D-2}^B\cos\alpha_{D-1}^B \\\sin\alpha_1^B...\sin\alpha_{D-2}^B\sin\alpha_{D-1}^B\end{pmatrix},\\
    \sigma\n=\sigma\begin{pmatrix}
    \cos\alpha_1 
    \\\sin\alpha_1\cos\alpha_2 \\\sin\alpha_1\sin\alpha_2\cos\alpha_3 \\\vdots  \\\sin\alpha_1...\sin\alpha_{D-2}\cos\alpha_{D-1} \\\sin\alpha_1...\sin\alpha_{D-2}\sin\alpha_{D-1}
    \end{pmatrix}    
\end{multline}

Here we set (by rotation) the direction of $\A$ to be the unit vector in the first dimension.
Thus, the cosine similarity between $\B$ and $\A$ is now the cosine value of the first angle ($\alpha^B_1$) of $\B$ in the spherical form (similarly for the noise vector $\n$). 
Note that the above parameterization does not change the entropy or conditional entropy since $\sigma\n$ is uniformly distributed in each direction.
Also, note that the scaling of a distribution shifts its entropy by a logarithm of the scaling factor. 
Thus, the first term is a constant given symmetry (randomness is in $\{\alpha_k\}_{k=1...D-1}$), i.e.,
\begin{equation}
    \ent(\sigma\n\cdot\B) = \ent^{\mathrm{proj}}(\mathbb{S}^{D-1}) + \log(\sigma_B\sigma)
    \label{eq:first-term}
\end{equation}
Next, we calculate $\ent(\sigma\n\cdot\B|\sigma\n\cdot\A)$. First, let's check $\ent(\sigma\n\cdot\B|\sigma\n\cdot\A=y)$, i.e., the entropy of $\sigma\n\cdot\B$ when the project of $\sigma\n$ on $\A$ is $y$. We have $y=\sigma_A\sigma\cos\alpha_1$, and
\begin{multline}
    \sigma\n\cdot\B=
    \sigma_B\cos\alpha_1^B\sigma\cos\alpha_1+
    \sigma_B\sin\alpha_1^B \sigma\sin\alpha_1\cdot\\ \begin{pmatrix}
    \cos\alpha_2^B \\\vdots \\\sin\alpha_2^B...\sin\alpha_{D-1}^B\end{pmatrix}^T
    \begin{pmatrix}
    \cos\alpha_2 \\\vdots \\\sin\alpha_2...\sin\alpha_{D-1}\end{pmatrix}\\
    = y\cos\alpha_1^B\dfrac{\sigma_B}{\sigma_A} +
    \sigma_B\sin\alpha_1^B\sigma\sqrt{1-(\dfrac{y}{\sigma_A\sigma})^2}\cdot\\
    \begin{pmatrix}
    \cos\alpha_2^B \\\vdots \\\sin\alpha_2^B...\sin\alpha_{D-1}^B\end{pmatrix}^T
    \begin{pmatrix}
    \cos\alpha_2 \\\vdots \\\sin\alpha_2...\sin\alpha_{D-1}\end{pmatrix}
\end{multline}
Thus, we have:
\begin{multline}
    \ent(\sigma\n\cdot\B|\sigma\n\cdot\A=y) = \ent^{\mathrm{proj}}(\mathbb{S}^{D-2}) +\\ \log(\sigma_B\sin\alpha_1^B\sigma\sqrt{1-(\dfrac{y}{\sigma_A\sigma})^2}) 
\end{multline}

Suppose that the probability density function of $y$ is $f_Y$, then we have:
\begin{multline}
    \ent(\sigma\n\cdot\B|\sigma\n\cdot\A) = 
    \int f_Y(y)\ent(\sigma\n\cdot\B|\sigma\n\cdot\A=y)\mathrm{d}y\\
    = \int f_Y(y)\ent^{\mathrm{proj}}(\mathbb{S}^{D-2})\mathrm{d}y +\\ \int f_Y(y)\log(\sigma_B\sin\alpha_1^B\sigma\sqrt{1-(\dfrac{y}{\sigma_A\sigma})^2})\mathrm{d}y\\
    = \ent^{\mathrm{proj}}(\mathbb{S}^{D-2}) + \log(\sigma_B\sin\alpha_1^B\sigma) +\\ \int_{-\sigma_A\sigma}^{\sigma_A\sigma} f_Y(y)\sqrt{1-(\dfrac{y}{\sigma_A\sigma})^2}\mathrm{d}y\\
    = \log(\sigma_B\sin\alpha_1^B\sigma) + \ent^{\mathrm{proj}}(\mathbb{S}^{D-2}) + \int_0^\pi f_{\alpha_1}(\alpha)\sin(\alpha)\mathrm{d}\alpha
    \label{eq:second-term}
\end{multline}

Combining Eq.~\eqref{eq:first-term} and Eq.~\eqref{eq:second-term}, and denote the last term in Eq.~\eqref{eq:second-term} as $h(f_{\alpha1})$ ({\it constant}) with $f_{\alpha_1}$ the probability density function of $\alpha_1$, then we have:
\begin{multline}
    \mi(\hat{I}(\p_i),\hat{I}(\p_j)) = \ent^{\mathrm{proj}}(\mathbb{S}^{D-1}) + \log(\sigma_B\sigma) -\\ \log(\sigma_B\sin\alpha_1^B\sigma) - \ent^{\mathrm{proj}}(\mathbb{S}^{D-2}) - h(f_{\alpha1})\\
    = \log(\dfrac{1}{\sin\alpha_1^B}) + \ent^{\mathrm{proj}}(\mathbb{S}^{D-1}) - \ent^{\mathrm{proj}}(\mathbb{S}^{D-2}) - h(f_{\alpha1})\\
    = \log(\dfrac{1}{\sqrt{1-\cos^2\alpha_1^B}}) + \ent^{\mathrm{proj}}(\mathbb{S}^{D-1}) - \ent^{\mathrm{proj}}(\mathbb{S}^{D-2}) - h(f_{\alpha1})\\
    (\propto \|\cos\alpha_1^B\|)
    \label{eq:mutual-info}
\end{multline}
According to Eq.~\eqref{eq:mutual-info}, we can see that the mutual information between the two pixels $\p_i,\p_j$ under the perturbation-induced joint distribution is positively correlated to the absolute value of the cosine similarity of their gradients with respect to the perturbed network weights.

\subsection{Implementation details}

\subsubsection{Training details}
\begin{figure}[!t]
\centering
  \includegraphics[width=\linewidth]{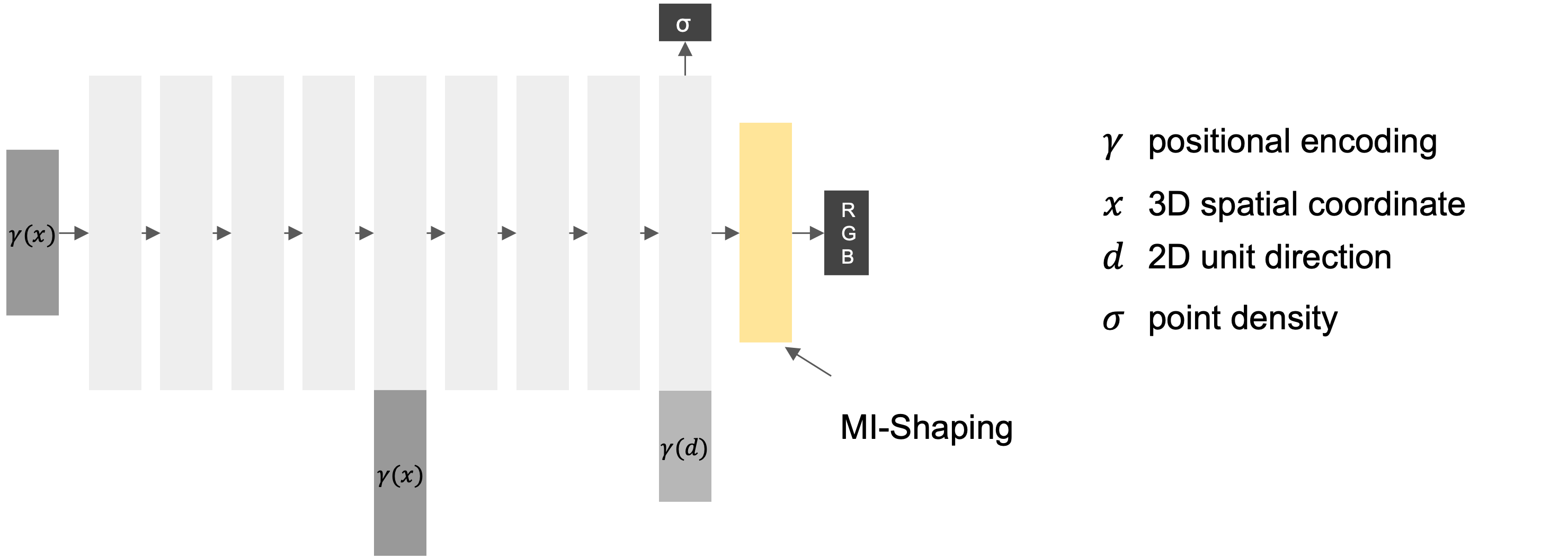}
  \caption{The NeRF architecture and the network parameters have been shaped. $\gamma$ denotes positional encoding, $x$ denotes 3D spatial coordinate, $d$ represents 2D unit direction, and $\sigma$ represents point density. The proposed mutual information shaping (MI-shaping) in our current implementation is applied on the $3\times 128$-dimensional parameters of NeRF's RGB linear layer, shown as the orange layer in the figure.}
  \label{fig:architecture}
\end{figure}

The MI-shaping process follows Sec.~\ref{sec:mig-shaping} in the main text, where we first train the NeRF with only images, and then shape it via the proposed contrastive loss on the gradients. The gradients we use to shape NeRF are computed with the gray values of pixels (mean value of the 3 RGB channels), with respect to the $3\times 128$-dimensional parameters of NeRF's RGB linear layer, as shown in Fig.~\ref{fig:architecture}. 

For each epoch, we randomly sample a batch of 64 rays across all the training views. For each sampled ray, we select from the other rays whose corresponding DINO feature similarity with the sampled ray is higher than a threshold as the positive sample, and the others with lower similarity as negative samples (the total number of samples in the contrastive loss of a certain ray is also 64). 
The threshold for each scene is adaptively adjusted during the training process. We first set an interval $[0.5, 0.8]$ for the threshold, then for each step, we subtract 0.001 from the threshold if the ratio of positive samples is lower than $5\%$, and add 0.001 to the threshold if the ratio is larger than $15\%$, unless the threshold exceeds the interval. The loss is shown in Eq.~\eqref{eq:shaping-loss}, where $\lambda=0.01, \gamma=0.01$, optimized by the Adam \cite{kingma2014adam} optimizer, with an initial learning rate of $5\times 10^{-4}$ for 10000 epochs.

The training runs for 24 hrs with color reconstruction and then 8 hrs with the shaping loss activated (mem: 1350 MB). Semantic-NeRF\cite{zhi2021place} and DINO-NeRF\cite{kobayashi2022distilledfeaturefields} take 28 hrs.

\subsubsection{Label propagation with JacobiNeRF in 3D}
Here we elaborate on the implementation details of the label propagation method denoted as J-NeRF 3D in the main text. 
The only difference between J-NeRF 3D and J-NeRF 2D is that we collect the perturbation responses of sampled points in 3D instead of 2D pixel difference, and then volume render them into segmentation logits for pixels. 
To render a 2D segmentation label, we first sample points along the rays following the hierarchical sampling strategy of NeRF \cite{mildenhall2021nerf}. Then we render the points' unperturbed radiance values and K (K is the number of classes) times the perturbed values, thus obtaining K-dimensional differences for each point. We integrate the differences of the sampled points along each ray following the volume render process in Eq.~\eqref{eq:volume_rend} to get a K dimensional segmentation logits in 2D pixel space.

\subsubsection{Label propagation under dense setting}
{\bf Adaptive gradient sampling.}
For efficiency, we have to choose representative gradients for each class to perform perturbation rather than perturbing along all the given labels' gradients. To avoid loss of label information, we employ an adaptive gradient sampling strategy to make the gradients as sufficient as possible to reconstruct the given dense label.

\begin{figure*}[!t]
  \centering
  \includegraphics[width=0.9\linewidth]{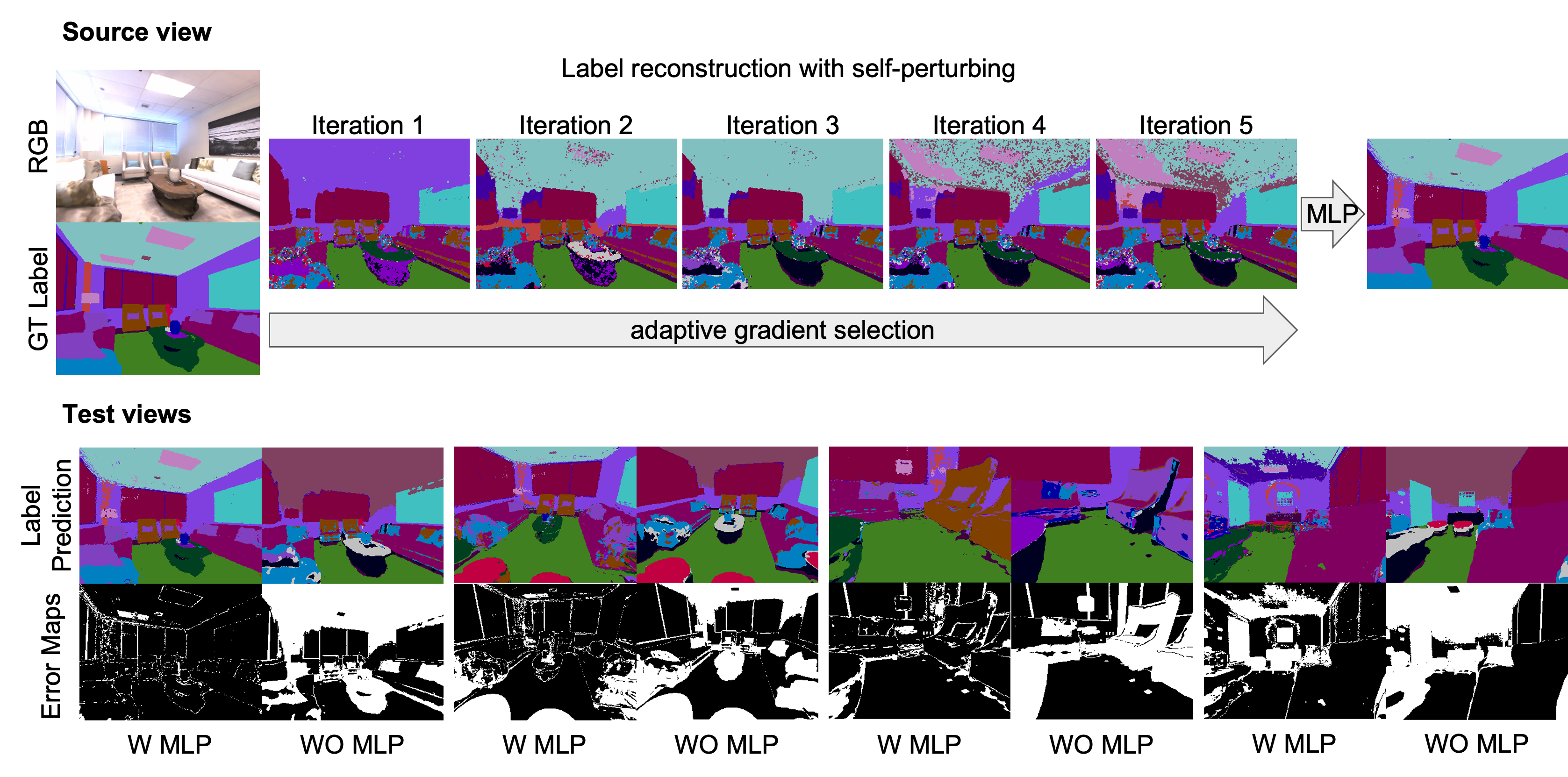}
  \caption{The effect of adaptive gradient sampling and aggregation MLP. Top: as more and more qualified gradients are selected, the reconstruction quality of the dense input label also gets better. Bottom: the denoising MLP can make better decode the perturbation responses.}
  \label{fig:propagation-dense}
\end{figure*}

For the first round, we randomly select 20 combinations of labeled pixels, each combination containing K pixels (K is the number of classes). 
We perturb along these gradients and get the response of the source view (whose dense label is known), thereby getting label prediction of the source view from the Argmax of the response. 
We evaluate the quality of the predicted label by calculating gain (mIoU) with the given dense label and choose the combination of gradients that yields the largest gain. 
For the next iteration, we sample another 20 combinations of gradients, and choose the one with the largest gain. We discard these gradients if the gain from these 20 combinations is not larger than zero. 
We repeat the above process until we get 5 qualified selections, i.e., each selection gives us K labels, in total 5$\times$K labels.
Fig.~\ref{fig:propagation-dense} shows how we adaptively sample representative gradients. As the iteration of adaptive sampling goes on, the selected gradients can better reconstruct the given dense label, indicating that the gradients encode more and more information about the dense label. 

{\bf Aggregation MLP.}
We leverage a lightweight aggregation/denoising MLP to further denoise the perturbed response with information from the dense label. The aggregation MLP maps from each pixel's K-dimensional (K is the number of classes) perturbation responses to K-dimensional segmentation logits. After we adaptively select the gradients and get the responses on the source view by perturbing along the selected gradients, we train the aggregation MLP from scratch under aggregation loss $L_{agg}$:
\begin{equation}
    L_{agg} = - \sum_{r\in \mathcal{R}}\sum_{k=1}^Kp^k(r)\log p^k_{agg}(r),
\end{equation}
where $\mathcal{R}$ are the rays within the source view, $p^k$ is the probability at class $k$ of the ground-truth label, and $p^k_{agg}$ is the probability at class $k$ predicted by the aggregation MLP from the perturbed response. $L_{agg}$ is a multi-class cross-entropy loss to encourage the denoised predicted label to be consistent with the given ground-truth dense labels.
The MLP has 3 linear layers and ReLU activation, with $K\times256+256\times128+128\times K$ parameters. 
The MLP is optimized with the Adam \cite{kingma2014adam} optimizer with an initial learning rate of $1\times10^{-3}$. We train it for 200000 iterations, approximately 30 minutes. On the right of Tab.~\ref{tab:abl_plain}, we show the quantitative effect of the lightweight aggregation MLP ({\it Please note that J-NeRF+ represents the propagation with the denoising MLP.}). Fig.~\ref{fig:propagation-dense} shows the denoising effect qualitatively.

\subsection{Additional results}

\subsubsection{Novel view synthesis quality}
We also report the novel view synthesis quality of the plain NeRF and the proposed JacobiNeRF in Tab.~\ref{tab:psnr}. 
As confirmed, the MI-shaping process does not degrade the performance of view synthesis, i.e., the image quality is similar to the one without shaping.

\begin{table}[h]
\centering
\setlength\tabcolsep{2pt}
\begin{tabular}{ccc}
    \hline
    Method & NeRF & J-NeRF\\
    \hline
    PSNR $\uparrow$ & 18.88 & 18.79\\
    \hline
\end{tabular}
\vspace{-0.1cm}
\caption{The performance of novel view synthesis on the Replica dataset measured by the peak signal-to-noise ratio (PSNR).}
\label{tab:psnr}
\end{table}

\subsubsection{Propagation capability of plain NeRF} To further validate the effectiveness of our NeRF shaping method, we replace JacobiNeRF with plain NeRF (w/o MI-shaping) in our propagation pipeline and propagate in 2D by perturbing along the gradients of labeled pixels. The semantic segmentation propagation results, in the sparse setting, averaged over the 7 scenes from Replica are shown in Tab.~\ref{tab:abl_plain} (left). As observed, J-NeRF 2D significantly outperforms plain NeRF, proving that our method shapes NeRF effectively to make it more suitable for information propagation.

\begin{table}[h]
\centering
\setlength\tabcolsep{1.1pt}
    \begin{tabular}{ccc}
    \toprule
    Method & NeRF & J-NeRF\\
    \midrule
    mIoU$\uparrow$ & 0.067 & \textbf{0.263}\\
    Avg Acc$\uparrow$ & 0.182 & \textbf{0.489}\\
    Total Acc$\uparrow$ & 0.199 & \textbf{0.483}\\
    \bottomrule
    \end{tabular}
    \quad
    \begin{tabular}{ccc}
    \toprule
    Method & J-NeRF & J-NeRF+\\
    \midrule
    mIoU$\uparrow$ & 0.411 & \textbf{0.524}\\
    Avg Acc$\uparrow$ & \textbf{0.724} & 0.689\\
    Total Acc$\uparrow$ & 0.633 & \textbf{0.864}\\
    \bottomrule
    \end{tabular}
    \vspace{-0.1cm}
    \caption{Effectiveness of the proposed shaping on a plain NeRF (left), and of the lightweight perturbation response decoder (right).}
\label{tab:abl_plain}
\end{table}

\subsubsection{Learnable ``argmax'' for dense label propagation}
We study the effectiveness of the denoising MLP used in the dense setting by comparing to those labels from the Argmax of the perturbation response. Results averaged over 7 scenes are shown in Tab.~\ref{tab:abl_plain} (right), indicating that the lightweight MLP improves the utilization of the dense labels.

\subsubsection{Distribution of selected regularization points} 
\begin{figure}[!t]
\centering
  \includegraphics[width=0.3\linewidth]{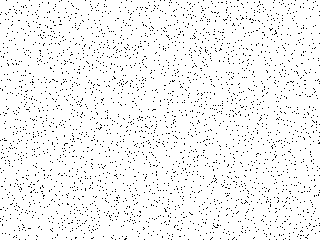}
    \vspace{-0.2cm}
    \caption{Distribution of selected regularization points.}
  \label{fig:visualize_sample}
\end{figure}

Points sampled during the contrastive training (MI-shaping) process are uniformly random as in Fig.~\ref{fig:visualize_sample}.

\subsubsection {Fewer regularization samples.}
\begin{figure}[!t]
\centering
  \includegraphics[width=\linewidth]{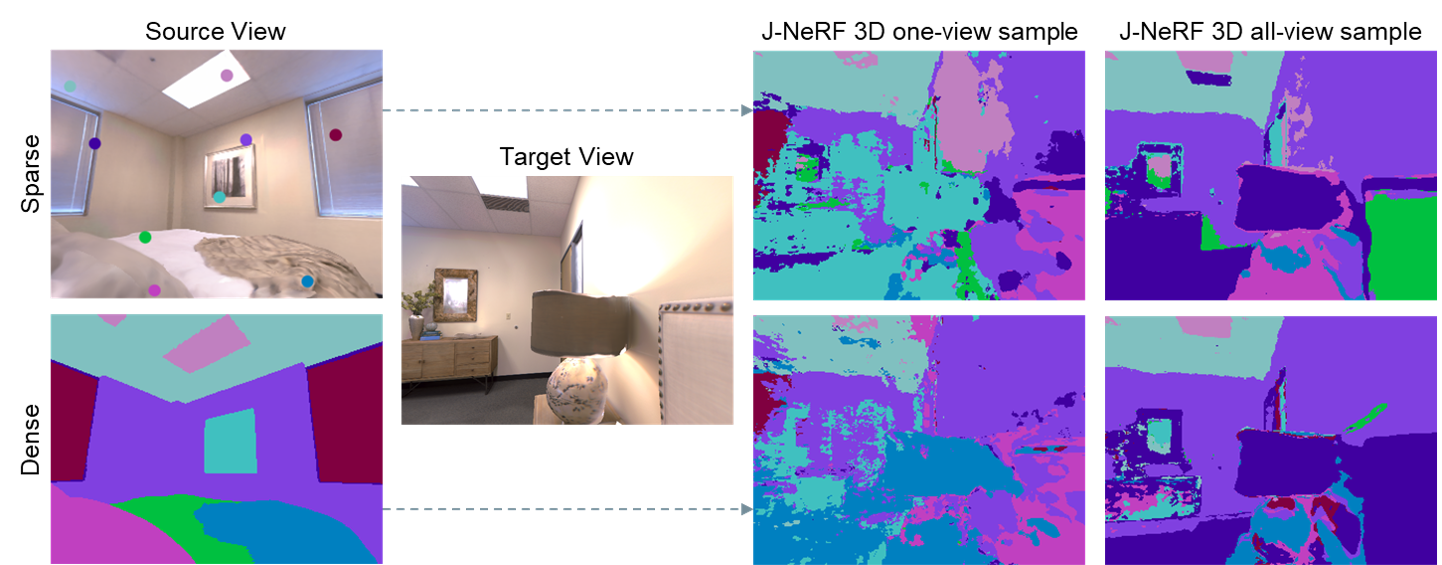}
  \vspace{-0.3cm}
  \caption{Qualitative comparison: one-view v. all-view regularization.}
  \vspace{-0.1cm}
  \label{fig:sample}
\end{figure}

In Fig.~\ref{fig:sample}, we observe that when the selected samples are constrained to one view, the clustering effect reduces compared to all-view regularization.

\subsection{Additional Visualization}
\begin{figure*}[!t]
  \centering
  \includegraphics[width=1.0\linewidth]{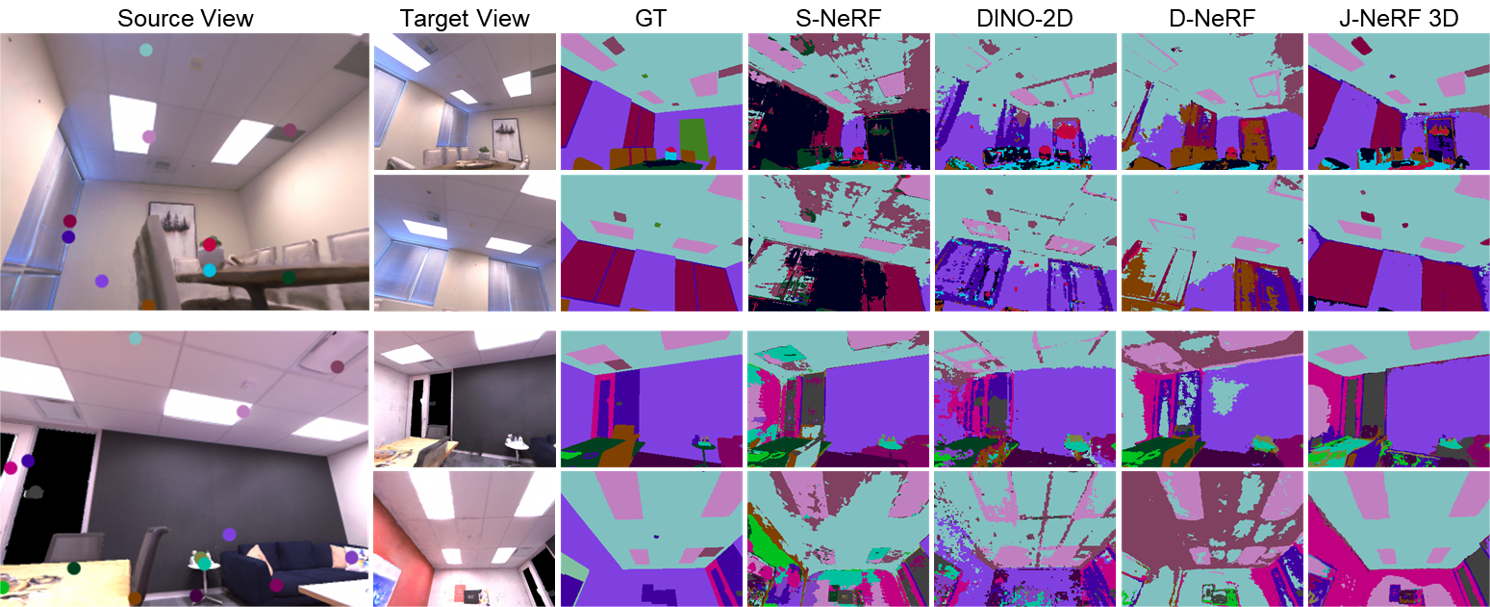}
  \vspace{-0.6cm}
  \caption{Qualitative results of semantic segmentation propagation under the sparse setting. Examples are from the Replica dataset.}
  \vspace{-0.2cm}
  \label{fig:replica_sparse}
\end{figure*}

\begin{figure*}[!t]
  \centering
  \includegraphics[width=1.0\linewidth]{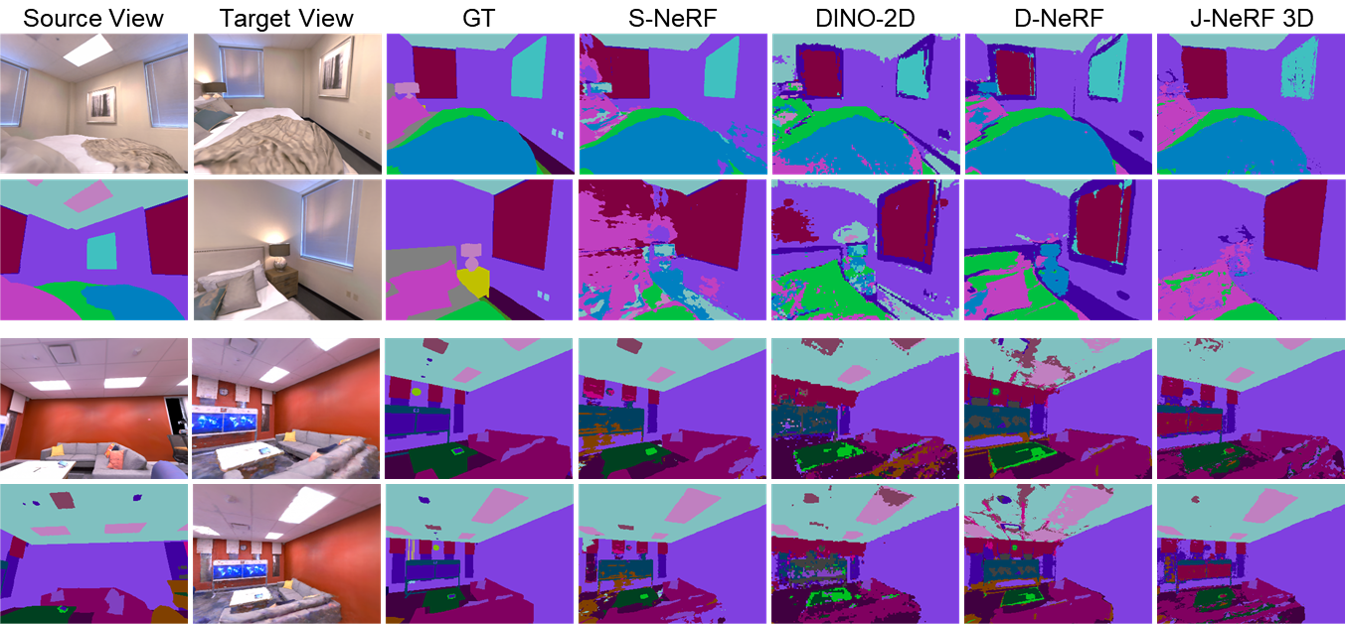}
  \vspace{-0.6cm}
  \caption{Qualitative results of semantic segmentation propagation under the dense setting. Examples are from the Replica dataset.}
  \vspace{-0.2cm}
  \label{fig:replica_dense}
\end{figure*}

\begin{figure*}[!t]
  \centering
  \includegraphics[width=1.0\linewidth]{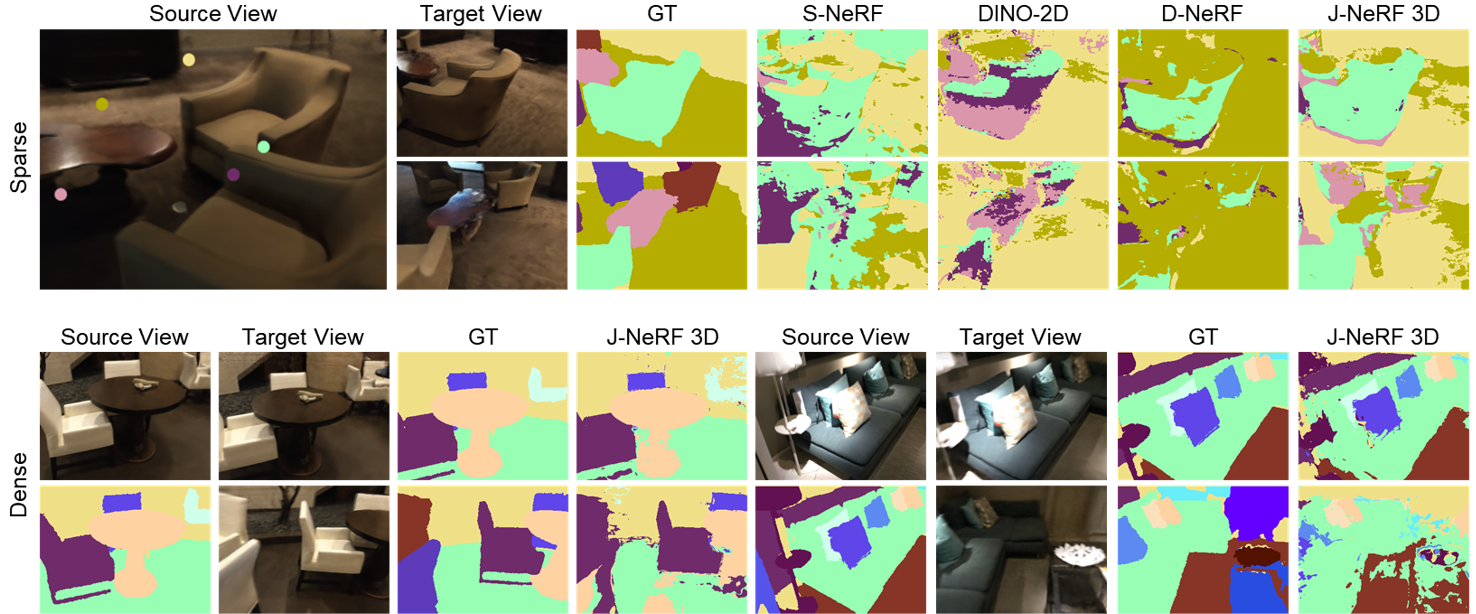}
  \vspace{-0.6cm}
  \caption{Qualitative results of instance segmentation propagation under both sparse and dense settings. Examples are from the ScanNet dataset.}
  \vspace{-0.2cm}
  \label{fig:scannet_ins}
\end{figure*}

Please see Fig.~\ref{fig:replica_sparse}, \ref{fig:replica_dense}, \ref{fig:scannet_ins} for more qualitative label propagation results.

%%%%%%%%% REFERENCES
\newpage
{\small
\bibliographystyle{ieee_fullname}
\bibliography{egbib}
}

\end{document}